%% file: main.tex
\pdfoutput=1

\documentclass[11pt]{article}
\usepackage{arabtex}

\usepackage{utf8}
\usepackage{emnlp2021}

\usepackage{times}

\usepackage[T1]{fontenc}

\usepackage[utf8]{inputenc}

\usepackage{latexsym}

\usepackage{microtype}

\usepackage{times}
\usepackage{url}
\usepackage{multirow}
\usepackage{makecell}
\usepackage{graphicx}
\usepackage{url}
\usepackage{paralist}
\usepackage{tabularx}
\usepackage{qtree}

\usepackage{subcaption}
\usepackage{wrapfig,lipsum,booktabs}

\usepackage{colortbl}
\usepackage{siunitx}
\usepackage{color,soul}

\usepackage{comment}

\usepackage{adjustbox}
\usepackage{appendix}
\usepackage{placeins}

\usepackage{tikz}
\usepackage{amsmath}
\usepackage{hyperref}
\usepackage{cleveref}

\crefname{section}{§}{§§}
\Crefname{section}{§}{§§}
\definecolor{light-gray}{gray}{0.85}

\usepackage{arydshln}

\usepackage{tikz}
\def\checkmark{\tikz\fill[scale=0.2](0,.35) -- (.25,0) -- (1,.7) -- (.25,.15) -- cycle;} 
\title{NADI 2022:\\The Third Nuanced Arabic Dialect Identification Shared Task}

\author{Muhammad Abdul-Mageed, Chiyu Zhang, AbdelRahim Elmadany, \\\textbf{Houda Bouamor},$^\dagger$ \textbf{Nizar Habash}$^\ddagger$\\
The University of British Columbia, Vancouver, Canada\\
$^\dagger$Carnegie Mellon University in Qatar, Qatar\\
$^\ddagger$New York University Abu Dhabi, UAE\\
  {\tt \{muhammad.mageed@,a.elmadany@,chiyuzh@mail\}.ubc.ca} \\
  {\tt ~~ hbouamor@cmu.edu ~~~ nizar.habash@nyu.edu}\\
  }

\begin{document}
\setcode{utf8}
\setarab 
\maketitle

\begin{abstract}
We describe the findings of the third Nuanced Arabic Dialect Identification Shared Task (NADI 2022). NADI aims at advancing state-of-the-art Arabic NLP, including Arabic dialects. It does so by affording diverse datasets and modeling opportunities in a standardized context where meaningful comparisons between models and approaches are possible. NADI 2022 targeted both dialect identification (Subtask 1) and dialectal sentiment analysis (Subtask 2) at the country level. A total of $41$ unique teams registered for the shared task, of whom $21$ teams have participated (with $105$ valid submissions). Among these, $19$ teams participated in Subtask 1, and $10$ participated in Subtask 2. The winning team achieved  F\textsubscript{1}=$27.06$ on Subtask~1 and F\textsubscript{1}=$75.16$ on Subtask~2, reflecting that the two subtasks remain challenging and motivating future work in this area. We describe the methods employed by the participating teams and offer an outlook for NADI. 
\end{abstract}

\input{intro}

\input{lit}
\input{tasks}

\input{datasets}
\input{results}

\input{conc}
\input{ack}

\bibliography{dlnlp,NADI2022-paper}
\bibliographystyle{acl_natbib}


\end{document}

%% file: intro.tex
\section{Introduction}\label{sec:intro}

\textit{Arabic} is a collection of languages and language varieties some of which are not mutually intelligible, although it is sometimes conflated as a single language. \textit{Classical Arabic (CA)} is the variety used in old Arabic poetry and the Qur'an, the Holy Book of Islam. CA continues to be used to date, side by side with other varieties, especially in religious and literary discourses. CA is also involved in code-switching contexts with \textit{Modern Standard Arabic (MSA)}~\cite{mageed2020microdialect}. In contrast, as its name suggests, MSA is a more modern variety~\cite{badawi1973levels} of Arabic. MSA is usually employed in pan-Arab media such as AlJazeera network and in government communication across the Arab world.\footnote{\url{https://www.aljazeera.com/}} \textit{Dialectal Arabic (DA)} is the term used to collectively refer to Arabic dialects. DA is sometimes defined regionally into categories such as Gulf, Levantine, Nile Basin, and North African~\cite{Habash:2010:introduction,mageed2015Dissert}. More recent treatments of DA focus on more nuanced variation at the country or even sub-country levels~\cite{Bouamor:2018:madar,mageed2020microdialect}. Many of the works on Arabic dialects thus far have focused on dialect identification, the task of automatically detecting the source variety of a given text or speech segment. 

In this paper, we introduce the findings and results of the third Nuanced Arabic Dialect Identification Shared Task (NADI 2022). NADI aims at encouraging research work on Arabic dialect processing by providing datasets and diverse modeling opportunities under a common evaluation setup. The first instance of the shared task, NADI 2020~\cite{mageed:2020:nadi}, focused on province-level dialects. NADI 2021~\cite{abdul-mageed-etal-2021-nadi}, the second iteration of NADI, focused on distinguishing both MSA and DA according to their geographical origin at the country level. NADI 2022 extends on both editions and offers a richer context as it targets both Arabic dialect identification and  \textit{and} dialectal sentiment analysis.

\begin{figure}[t]
  \begin{center}
  \frame{\includegraphics[width=\linewidth]{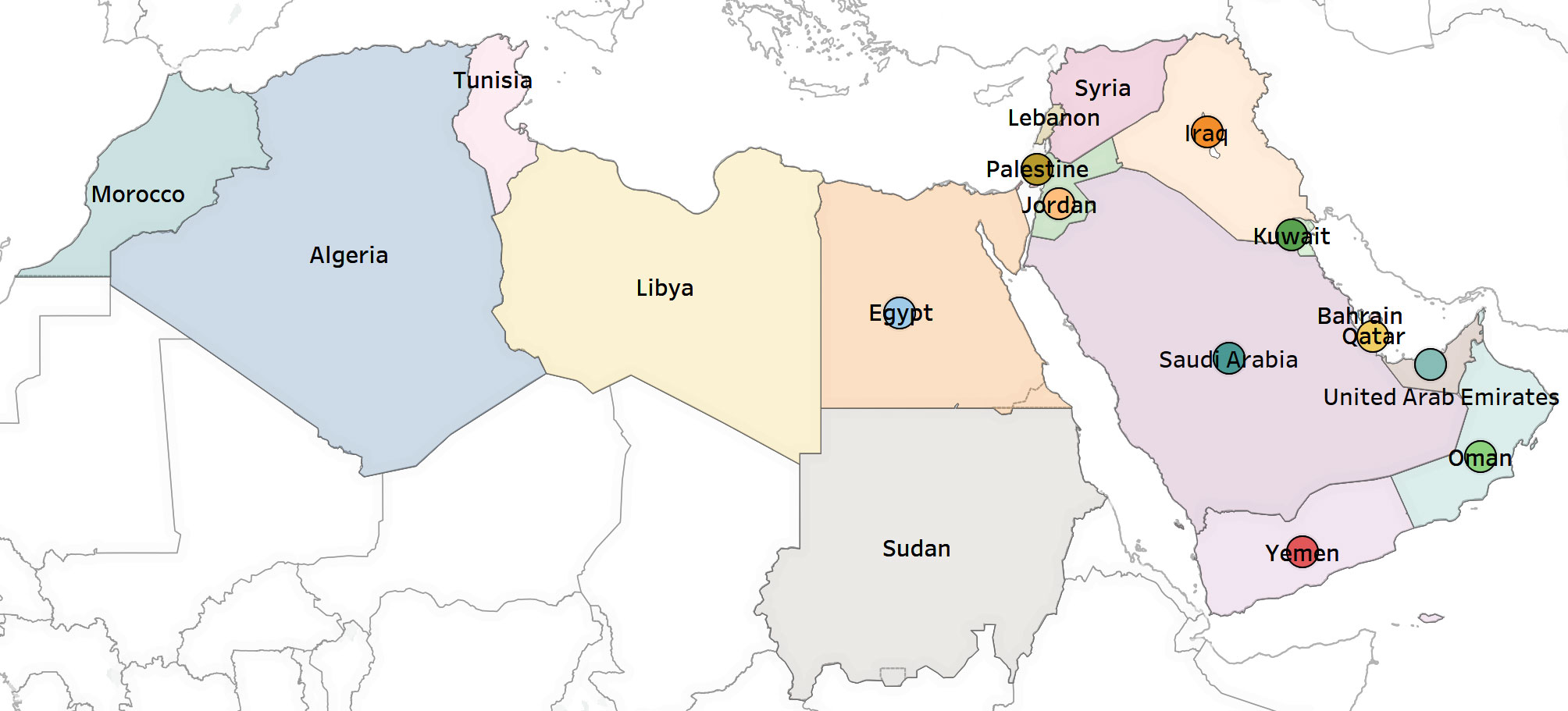}}
  \end{center}
\caption{A map of the Arab World showing the $18$ countries in the \textit{Subtask~1} dataset and the $10$ countries in the \textit{Subtask~2} dataset. Each country is coded in a color different from neighboring countries. Subtask~2 countries are coded as circles with dark color.}
\label{fig:map}
\end{figure}

NADI 2022 shared tasks proposes two subtasks: \textbf{Subtask~1} on dialect identification, and  \textbf{Subtask~2} on dialect sentiment analysis. While we invited participation in either of the two subtasks, we encouraged teams to submit systems to \textit{both} subtasks. By offering two subtasks, our hope was to receive systems that exploit diverse machine learning and other methods and architectures such as multi-task learning systems, ensemble methods, sequence-to-sequence architectures in single models such as the text-to-text Transformer, etc. Many of the submitted systems investigated diverse approaches, thus fulfilling our objective. 

A total of $41$ unique teams registered for NADI 2022. Of these, $21$ unique teams actually made submissions to our leaderboard (n=$105$ valid submissions). We received $16$ papers from $15$ teams, of which we accepted $15$ for publication. Results from participating teams show that both dialect identification at the country level and dialectal sentiment analysis from short sequences of text remain challenging even to complex neural methods. These findings clearly motivate future work on both tasks.

The rest of the paper is organized as follows: Section~\ref{sec:lit} provides a brief overview of Arabic dialect identification and sentiment analysis. We describe the two subtasks and NADI 2022 restrictions in Section~\ref{sec:tasks}. Section~\ref{sec:data_eval} introduces shared task datasets and evaluation setup. We present participating teams and shared task results and provide
a high-level description of submitted systems in Section~\ref{sec:results}. We conclude in Section~\ref{sec:conc}.

%% file: lit.tex
\section{Literature Review}\label{sec:lit}
\subsection{Arabic Dialects} 
Arabic can be categorized into CA, MSA, and DA. Although CA and MSA have been studied extensively~\cite{Harrell:1962:short,Cowell:1964:reference,badawi1973levels,Brustad:2000:syntax,Holes:2004:modern}, DA is has received more attention only in recent years. One major challenge for studying DA has been the lack of resources. For this reason, most pioneering DA works focused on creating resources, usually for only a small number of regions or countries~\cite{Gadalla:1997:callhome,diab2010colaba,al2012yadac,sadat2014automatic,Smaili:2014:building,Jarrar:2016:curras,Khalifa:2016:large,Al-Twairesh:2018:suar,el-haj-2020-habibi}. A number of works introducing multi-dialectal datasets and regional level detection models followed~\cite{zaidan2011arabic,Elfardy:2014:aida,Bouamor:2014:multidialectal,Meftouh:2015:machine}. 
 
Some of the earliest Arabic dialect identification shared tasks were offered as part of the VarDial workshop. These shared tasks used speech broadcast transcriptions~\cite{malmasi2016discriminating}, and later integrated acoustic features~\cite{zampieri2017findings} and phonetic features~\cite{zampieri2018language} extracted from raw audio.

The Multi-Arabic Dialects Application and Resources (MADAR) project~\cite{Bouamor:2018:madar} was the first that introduced finer-grained dialectal data and a lexicon. The MADAR data was used for dialect identification at the country and city levels covering $25$ cities in the Arab world \cite{Salameh:2018:fine-grained,obeid-etal-2019-adida}. The MADAR data was commissioned rather than being naturally occurring, which might not be the best for dialect identification, especially when considering dialect identification in the social media context. Several larger datasets covering $10$-$21$ countries were then introduced~\cite{Mubarak:2014:using,Abdul-Mageed:2018:you,Zaghouani:2018:araptweet, abdelali-etal-2021-qadi,issa-etal-2021-country,baimukan2022hierarchical}. These datasets were mainly compiled from naturally-occurring posts on social media platforms such as Twitter. Some approaches for collecting dialectal data are unsupervised. A recent example is~\newcite{althobaiti2022creation} who describe an approach for automatically tagging Twitter posts with $15$ country-level dialects and extracting relevant word lists. Some works also gather data at the fine-grained level of cities. For example, \citet{mageed2020microdialect} introduced a Twitter dataset and a number of models to identify country, province, and city level variation in Arabic dialects. The NADI shared task~\cite{mageed:2020:nadi,abdul-mageed-etal-2021-nadi} built on these efforts by providing datasets and common evaluation settings for identifying Arabic dialects. \newcite{althobaiti2020automatic} is a relatively recent survey of computational work on Arabic dialects.

\subsection{Sentiment Analysis}
Besides dialect identification, several studies investigate socio-pragmatic meaning (SM) exploiting Arabic data. SM refers to intended meaning in real-world communication and how utterances should be interpreted within the social context in which they are produced~\cite{thomas2014meaning,zhang2022decay}. Typical SM tasks include sentiment analysis~\cite{Abdul-Mageed:2014:samar,mageed2019modeling}, emotion recognition~\cite{alhuzali2018enabling}, age and gender identification~\cite{abbes2020daict}, offensive language detection~\cite{mubarak2020overview,elmadany2020leveraging}, and sarcasm detection~\cite{farha2020arabic}. In NADI 2022, we focus on sentiment analysis of Arabic dialects in social media. Several studies of Arabic sentiment analysis are listed in surveys such as \newcite{elnagar2021sentiment} and \newcite{alhumoud2021arabic}. Most of these studies target sentiment in MSA. Recently, there are some studies that target sentiment in Arabic dialects in social media sources such Twitter. Some of these studies create datasets~\cite{guellil2020arautosenti, al2021arasencorpus,abo2021multi,alowisheq2021marsa, hassan2021asad,alwakid2022muldasa}, focusing on one or more dialects or regions~\cite{mageed2020aranet,fourati2020tunizi, guellil2020arabic, almuqren2021aracust, guellil2021semi,farha2021benchmarking, a2022sentiment}. Many of the previous sentiment analysis works, however, either do not distinguish dialects altogether or focus only on a few dialects such as Egyptian, Levantine, or Tunisian. This motivates us to introduce the dialectal sentiment analysis subtask as part of NADI 2022.

To the best of our knowledge, our work is the first to enable investigating sentiment analysis in $10$ Arabic dialects. For our sentiment analysis subtask, we also annotate and release a novel dataset and facilitate comparisons in a standardized experimental setting. 

\subsection{The NADI Shared Tasks}

\paragraph{NADI 2020} The first NADI shared task,~\cite{mageed:2020:nadi} was co-located with the fifth Arabic Natural Language Processing Workshop (WANLP 2020)~\cite{wanlp-2020-arabic}. NADI 2020 targeted both country- and province-level dialects. It covered a total of $100$ provinces from $21$ Arab countries, with data collected from Twitter. It was the first shared task to target naturally occurring fine-grained dialectal text at the sub-country level. 

\paragraph{NADI 2021} The second edition of the shared task \cite{abdul-mageed-etal-2021-nadi} was co-located with WANLP 2021~\cite{wanlp-2021-arabic}. It targeted the same $21$ Arab countries and $100$ corresponding provinces as NADI 2020, also exploiting Twitter data. NADI 2021 improved over NADI 2020 in that non-Arabic data were removed. In addition, NADI-2021 teased apart the data into MSA and DA and focused on classifying MSA and DA tweets into the countries and provinces from which they are collected. As such, NADI 2021 had four subtasks: MSA-country, DA-country, MSA-province, and DA-province. 

\paragraph{NADI 2022} As introduced earlier, \textbf{this current edition} of NADI focuses on studying Arabic dialects at the country level as well as dialectal sentiment (i.e., sentiment analysis of data tagged with dialect labels). Our objective is that NADI 2022 can support exploring variation in social geographical regions that have not been studied before. 
We discuss NADI 2022 in more detail in the next section.

It is worth noting that NADI shared task datasets  are starting to be used for various types of (e.g., linguistic) studies of Arabic dialects, For example,~\newcite{alsudais2022similarities} studies the effect of geographic proximity on Arabic dialects exploiting datasets from MADAR~\cite{Bouamor:2018:madar} and NADI~\cite{mageed:2020:nadi,abdul-mageed-etal-2021-nadi}.



%% file: tasks.tex
\section{Task Description}\label{sec:tasks}

\subsection{Shared Task Subtasks}

The NADI 2022 shared task consists of two subtasks, both focused on dialectal Arabic at the country level. \textbf{Subtask~1} is about  dialect identification and \textbf{Subtask~2} is about sentiment analysis of Arabic dialects. We now introduce each subtask.

\paragraph{Subtask~1 (Dialect Identification)} The goal of Subtask~1 is to identify the specific country-level dialect of a given Arabic tweet. For this subtask, we reuse the training, development, and test datasets of $18$ countries from NADI 2021~\cite{abdul-mageed-etal-2021-nadi}. In addition to the test set of NADI 2021, we introduce a \textit{new} test set manually annotated with $k$ country-level dialects, where $k=10$ but is kept unknown to teams. We ask participants to submit system runs on these two test sets.

\paragraph{Subtask~2 (Dialectal Sentiment Analysis)} The goal of Subtask~2 is to identify the sentiment of a given tweet written in Arabic. Tweets are collected from $10$ different countries during the year of $2018$ and involve both MSA and DA. The data are manually labeled with sentiment tags from the set \{\textit{positive}, \textit{negative}, \textit{neutral}\}. More information about our data splits and evaluation settings for both Subtask~1 and Subtask~2 is given in Section~\ref{sec:data_eval}.

Figure~\ref{fig:map} shows the countries covered in NADI 2022 for both subtasks. 

\subsection{Shared Task Restrictions}
We follow the same general approach to managing the shared task we adopted in NADI 2020 and NADI 2021. This includes providing participating teams with a set of restrictions that apply to all subtasks, and clear evaluation metrics. The purpose of our restrictions is to ensure fair comparisons and common experimental conditions. In addition, similar to NADI 2020 and 2021, our data release strategy and our evaluation setup through the CodaLab online platform facilitated competition management, enhanced timeliness of acquiring results upon system submission, and guaranteed ultimate transparency. Once a team registered in the shared task, we directly provided the registering member with the data via a private download link. We provided the data in the form of the actual tweets posted to the Twitter platform, rather than tweet IDs. This guaranteed comparison between systems exploiting identical data. 

For both subtasks, we provided clear instructions requiring participants not to use any external data. That is, teams were required to only use the data we provided to develop their systems and no other datasets regardless how these are acquired. For example, we requested that teams do not search nor depend on any additional user-level information such as geolocation. To alleviate these strict constraints and encourage creative use of diverse (machine learning) methods in system development, we provided an unlabeled dataset of $10$M tweets in the form of tweet IDs. This dataset is provided in addition to our labeled Train and Dev splits for the two subtasks. To facilitate acquisition of this unlabeled dataset, we also provided a simple script that can be used to collect the tweets. We encouraged participants to use the $10$M unlabeled tweets in whatever way they wished. 


%% file: datasets.tex
\section{Shared Task Datasets and Evaluation}
\label{sec:data_eval}

\paragraph{TWT-10}  We collected $\sim10K$ tweets covering $10$ Arab countries (
\textit{Egypt}, \textit{Iraq}, \textit{Jordan}, \textit{KSA}, \textit{Kuwait}, \textit{Oman}, \textit{Palestine}, \textit{Qatar}, \textit{UAE}, and \textit{Yemen}) via the Twitter API.\footnote{\url{https://developer.twitter.com/en/docs/twitter-api}} The tweets were collected during the year of $2018$. We asked a total of three college-educated Arabic native speakers to annotate these tweets with three types of information: (1) \textit{dialectness} (MSA vs. DA), (2) \textit{$10$-way country-level dialects}, and (3) \textit{three-way sentiment labels} (i.e., \{$positive$, $negative$, $neutral$\}).  For each of the $10$ countries, $500$ tweets were labeled by two different annotators. We calculated the inter-annotator agreement using Cohen's Kappa . We obtained a Kappa (\textit{K}) of $0.85$ for the sentiment labeling task and \textit{K} of $0.41$ for the $10$-way dialect identification one. 
Table~\ref{tab:manual_anon_distri} also presents the distribution of dialect and sentiment classes. 
It also shows that MSA comprises $50.86\%$ of TWT-10 (while DA is $49.14\%$). 
Table~\ref{tab:samples} shows tweet examples with sentiment labels randomly selected from a number of countries representing different regions in our annotated dataset.

\input{tables/subtask2_distribution}

\input{tables/samples}

\paragraph{Subtask~1 (Dialect Identification)} We use the dataset of Subtask 1.2 of NADI 2021 (i.e., country-level DA)~\cite{abdul-mageed-etal-2021-nadi}. This dataset was collected using tweets covering $21$ Arab countries during a period of $10$ months (Jan. to Oct.) during the year of $2019$. It was heuristically labelled exploiting the users' geo-location feature and mobility patterns and automatically cleaned to exclude non-Arabic and MSA tweets. 
For the purpose of this shared task, we keep the same training, development, and test splits as NADI 2021 but we exclude data from Djibouti, Somalia, and Mauritania since these are poorly represented in the dataset. We call the resulting dataset \textbf{TWT-GEO}. TWT-GEO includes $18$ country-level dialects, split into \textbf{Train} ($\sim 20K$ tweets), \textbf{Dev} ($\sim 5K$ tweets), and \textbf{Test-A} ($\sim 4.8K$ tweets). 
We refer to the test set of TWT-GEO as Test-A since we use an additional test split for evaluation, \textbf{Test-B}. Test-B contains $1.5$K dialect tweets randomly sampled from the TWT-10 dataset described earlier. 
Table~\ref{tab:subtask1_distri} presents the class distributions in Subtask~1 Train, Dev, and Test splits (Test-A and Test-B).  


\input{tables/subtask_1_distribution}

\begin{figure}[t]
  \centering
  \includegraphics[width=\linewidth]{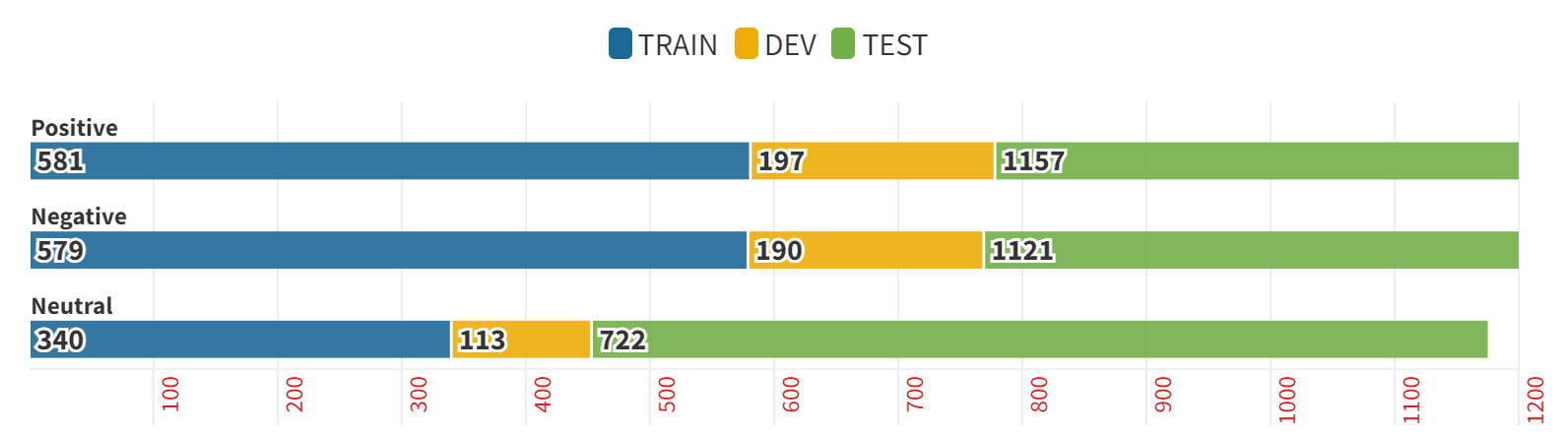}
\caption{Subtask~2 class distributions across data splits.}
\label{fig:subtask2_dustrib} 
\end{figure}

\paragraph{Subtask~2 (Sentiment Analysis)}  

 For this subtask, we use the manually annotated 5,000 tweets (including both MSA and dialects) in TWT-10. We randomly split the tweets into \textbf{Train} (1,500 tweets), \textbf{Dev} (500 tweets), and \textbf{Test} (3,000 tweets). We intentionally provide a small training dataset to encourage various approaches (e.g., \textit{few-shot} learning). Figure~\ref{fig:subtask2_dustrib} shows the distribution of sentiment classes across the data splits.

\input{tables/participant}

\paragraph{Unlabeled Dataset} We provide participants with a total of 10M unlabeled Arabic tweets in the form of tweet IDs. We refer to this collection as \textbf{UNLABELED-10M}. We collected these tweets in 2019. In UNLABELED-10M, Arabic was identified using Twitter language tag (\texttt{ar}). We included in our data package released to participants a simple script to collect these tweets. Participants were free to use UNLABELED-10M for any of the two subtasks.\footnote{Datasets for all the subtasks and UNLABELED-10M are available at \url{https://github.com/UBC-NLP/nadi}. More details about the data format can be found in the accompanying \texttt{README} file.} 

\paragraph{Evaluation Metrics} The official evaluation metric for \textbf{Subtask~1} is Macro-Averaged $F_1$-score. We evaluate on Test-A and Test-B separately, and use the average score between these two test sets as the final score of Subtask~1. For \textbf{Subtask~2}, $F\textsubscript{NP}$-score is the official metric, where we use the average of the $F_1$ scores of the \textit{positive} and \textit{negative} classes only while neglecting the neutral class. These metrics are obtained on blind test sets. We also report performance in terms of \textit{macro-averaged precision}, \textit{macro-averaged recall} and \textit{accuracy} for systems submitted to each of the two subtasks.

Each participating team was allowed to submit up to five runs for each test set of a given subtask, and only the highest scoring run was kept for each team. Although official results are based only on a blind test set, we also asked participants to report their results on the Dev sets in their papers. We set up two CodaLab competitions for scoring participant systems.\footnote{The different 
CodaLab competitions are available at the following links:  \href{https://codalab.lisn.upsaclay.fr/competitions/6514?secret_key=ce3736d6-03f2-4454-977c-1c88b7ef4d53}{Subtask~1}; \href{https://codalab.lisn.upsaclay.fr/competitions/6522?secret_key=86ae5871-9b85-4b1c-8e3d-129cd06be118}{ Subtask~2}.
} 
We plan to keep the Codalab competition for each subtask live post competition for researchers who would be interested in training models and evaluating their systems using the shared task blind test sets. For this reason, we will not release labels for the test sets of any of the subtasks. 

%% file: tables/subtask2_distribution.tex
\begin{table}[t!]
\centering
\begin{small}
\setlength{\tabcolsep}{4pt}
\begin{tabular}{lcccccc}

\toprule
\multirow{2}{*}{\textbf{Country}} & \multicolumn{2}{c}{\textbf{Dialect}} & \multicolumn{3}{c}{\textbf{Sentiment}}                                 & \multirow{2}{*}{\textbf{Total}} \\  \cmidrule(l){2-3} \cmidrule(l){4-6}
                                  & \textbf{MSA}     & \textbf{DA}     & \textbf{Pos} & \textbf{Neg} & \textbf{Neut} &

 \\
 \midrule
Egypt                                                  & 137                                    & 363                                                                       & 176                                       & 187                                       & 137                                      & 500                                    \\
Iraq                                                   & 314                                    & 186                                    & 230                                       & 219                                       & 51                                       & 500                                    \\
Jordan                                                 & 257                                    & 243                                     & 169                                       & 253                                       & 78                                       & 500                                    \\
KSA                                                    & 300                                    & 200                                   & 194                                       & 152                                       & 154                                      & 500                                    \\
Kuwait                                                 & 170                                    & 330                                    & 203                                       & 227                                       &70                                       & 500                                    \\
Oman                                                   & 340                                    & 160                                  & 166                                       & 179                                       & 155                                      & 500                                    \\
Palestine                                              & 248                                    & 252                                     & 159                                       & 169                                       & 172                                      & 500                                    \\
Qatar                                                  & 181                                    & 319                                   & 288                                       & 194                                       & 18                                       & 500                                    \\
UAE                                                    & 270                                    & 230                                    & 232                                       & 112                                       & 156                                      & 500                                    \\
Yemen                                                  & 326                                    & 174                                   & 118                                       & 198                                       & 184                                      & 500                                    \\
\midrule
                                                                             \textbf{Total} & \bf2,543& \bf2,457 &  \bf1,935  & \bf1,890  & \bf1,175   & \bf5,000\\\bottomrule
\end{tabular}%
\end{small}
\caption{The TWT-10 dataset class distributions.}
\label{tab:manual_anon_distri}
\end{table}

%% file: tables/samples.tex
\begin{table*}[ht]
\resizebox{\textwidth}{!}{%
\begin{tabular}{lrc}
\toprule
\multicolumn{1}{c}{\textbf{Country}} & \multicolumn{1}{c}{\textbf{Example}}                                                                                                    & \multicolumn{1}{c}{\textbf{Sentiment}} \\ \midrule
\multirow{3}{*}{Egypt}                 &  <محدش عنده فكة وجع !!>                                                                                                          & Negative                                \\ \cdashline{2-3}
   & <مرتضى : معروف و كوفي ملهمش مليم عندي و من يمتلك عرضا يغور>                                                                              & Neutral                                 \\\cdashline{2-3} 
    & < كوباية شاي خيالية لدرجة أني عايز آكل الكوباية الفاضية تسلم أيدي اصلن>                                                       & Positive                                \\ \hline

\multirow{3}{*}{Jordan}                & 
<وييننننك يا زووجي توخدني عبارريس>                                                                                           & Negative                                \\ \cdashline{2-3} 
& <منرجع منحكي: التعداد السكاني والمساكن، بدها تعرف الدولة وين ساكن>                                                                       & Neutral                                 \\ \cdashline{2-3} 
      & <شفتو حق جلسات وناسة وش سوا هههههههههاااااااي >                                                                                 & Positive                                \\ \hline
\multirow{3}{*}{KSA}                   & <لما تبي تحاكي احد بس ماتدري شتقول>                                                                                                    & Negative                                \\ \cdashline{2-3} 
             & <مادري ليش آذآ شفت احد نآيم احسه مسكين !!>                                                                                               & Neutral                                 \\ \cdashline{2-3} 
      & <الفجر يقيم خمس وعشر عشر ساعات يمديك توصل الشرقية تغطّ رجولك بالبحر وتتعشا وترجع >     & Positive                                \\ \hline
\multirow{3}{*}{Kuwait}                & <كالعاده الخشوف طاحسين بالتايم والسحوت يرضعون.>                                                                                          & Negative                                \\ \cdashline{2-3} 
& <احدٍ اذا مديتله طيبك .. يغيب ماتملي عيونه كبار الفعايل واحدٍ يموت إن كان سويت به طيب يخاف مايقدر يرد الجمايل>            & Neutral                                 \\ \cdashline{2-3} 
  & <اي والله أغليه و نفسي فيه مفتونه.>                                                                                           & Positive                                \\ \hline
\multirow{3}{*}{UAE}                   & 
\begin{tabular}[c]{@{}r@{}}
<ذا كنت تحبني صج ومن كل قلبك تعشقني ، حسسني بهالشي خلني أحس اني اهمك صج>\\
< مابي كم كلمه حلوه وانتا طول يومك مو معاي ومشغول بغيري>
\end{tabular}
 & Negative                                \\ \cdashline{2-3} 
                                      & < اقولك تدري منو سوالي فولو>                                                                                                         & Neutral                                 \\ \cdashline{2-3} 
                                      & < يبيله احلى بوسه على الصبح و لحس الخبق>                                                                                             & Positive                                
\\ \bottomrule
\end{tabular}%
}
\caption{Randomly picked dialectal tweets from select countries in our annotated data for Subtask 2.}\label{tab:samples}
\end{table*}

%% file: tables/subtask_1_distribution.tex

\begin{table}[t!]
\small
\centering
\setlength{\tabcolsep}{4pt}
\begin{tabular}{lrrrr}
\toprule
\textbf{Country} & \textbf{TRAIN} & \textbf{DEV} & \textbf{TEST-A}  & \textbf{TEST-B}\\ \midrule
Algeria          & 1,809           & 430          & 379    & -----         \\ 

Bahrain          & 215            & 52           & 50      & -----        \\ 

Egypt            & 4,283           & 1,041         & 1,025   &219         \\ 

Iraq             & 2,729           & 664          & 648    &117         \\ 

Jordan           & 429            & 104          & 101   & 144          \\ 

KSA              & 2,140           & 520          & 501  & 116           \\ 

Kuwait           & 429            & 105          & 103   &202          \\ 

Lebanon          & 644            & 157          & 119    & -----         \\ 

Libya            & 1,286           & 314          & 309    & -----         \\ 

Morocco          & 8,58            & 207          & 210   & -----          \\ 

Oman             & 1,501           & 355          & 360   &91          \\ 

Palestine        & 428            & 104          & 99    &160          \\ 

Qatar            & 215            & 52           & 51       &190       \\ 

Sudan            & 215            & 53           & 53     & -----         \\ 

Syria            & 1,287           & 278          & 279   & -----          \\ 

Tunisia          & 859            & 173          & 211   & -----          \\ 

UAE              & 642            & 157          & 157     &136        \\ 

Yemen            & 429            & 105          & 103   &99          \\ \bottomrule
\end{tabular}%
\caption{Distribution of classes for Subtask~1 data.}
\label{tab:subtask1_distri}
\end{table}

%% file: tables/participant.tex
\begin{table*}[t]
\footnotesize
\centering
\begin{tabular}{@{}llr@{}}
\toprule
\textbf{Team}                & \textbf{Affiliation}                                            & \textbf{Tasks} \\ \midrule
\textbf{259}~\cite{qaddoumi-2022-arabic}                 & New York University, USA                                             & 1              \\
\textbf{Ahmed and Khalil}~\cite{elshangiti-2022-ahmed}    & Independent Researcher, Morocco                                 & 1, 2           \\
\textbf{ANLP-RG}~\cite{fsih-2022-benchmarking} & \begin{tabular}[c]{@{}l@{}}Faculty of Economics and Management of Sfax, Tunisia \end{tabular}            & 2              \\
\textbf{BFCAI}~\cite{sobhy-2022-word}               & Benha University, Egypt                                         & 1              \\
\textbf{BhamNLP}             & \begin{tabular}[c]{@{}l@{}}King Abdulaziz University, KSA and Uni. of Birmingahm, UK \end{tabular} & 2              \\
\textbf{Elyadata}           & ELYADATA, Tunisia                                               & 1              \\
\textbf{Giyaseddin}~\cite{bayrak-2022-domain}          & Marmara University, Turkey                                      & 1, 2           \\
\textbf{GOF}~\cite{jamal-2022-arabic}                 & University of Windsor, Canada                                   & 1              \\
\textbf{iCompass}~\cite{messaoudi-2022-icompass}            & iCompass, Tunisia                                               & 1              \\
\textbf{ISL-AAST}            & Arab academy for science and technology, Egypt                  & 1, 2           \\
\textbf{MTU\_FIZ}~\cite{shammary-2022-tfidf}            & Munster Technological University, Ireland                       & 1              \\
\textbf{NLP\_DI}~\cite{kanjirangat-2022-nlpdi}             & Dalle Molle Institute for AI, Switzerland  & 1              \\
\textbf{Oscar\_Garibo}       & Valencian International University, Spain                       & 1, 2           \\
\textbf{Pythoneers}~\cite{attieh-2022-arabic}          & Aalto University, Finland                                       & 1, 2           \\
\textbf{rematchka}~\cite{abdelsalam2022dialect}           & Cairo University, Egypt                                         & 1, 2           \\
\textbf{RUTeam}              & Reichman University, Israel                                     & 1, 2           \\
\textbf{SQU}~\cite{aalabdulsalam-2022-squcs}                 & Sultan Qaboos University, Oman                                  & 1              \\
\textbf{SUKI}~\cite{jauhiainen-2022-optimizing}                & University of Helsinki, Finland                                 & 1              \\
\textbf{UniManc}~\cite{Khered-2022-building}             & The University of Manchester, UK                                & 1, 2           \\
\textbf{XY}~\cite{alshenaifi-2022-arabic}                  & Kind Saud University, KSA                                       & 1              \\
\textbf{zTeam}               & British University in Dubai, UAE                                & 1              \\ \bottomrule
\end{tabular}
\caption{List of teams that participated in either one or the two of subtasks. Teams with accepted papers are cited.}\label{tab:teams}
\end{table*}

%% file: results.tex
\section{Shared Task Teams \& Results}\label{sec:results}

\subsection{Participating Teams}
\label{sec:teams}

We received a total of $41$ unique team registrations. After the testing phase, we received a total of $105$ valid submissions from $21$ unique teams. The breakdown across the subtasks is as follows: $42$ submission for Test-A of Subtask~1 from $19$ teams, $41$ submissions for Test-B of Subtask~1 from $19$ teams, $22$ submissions for Subtask~2 from $10$ teams. Table~\ref{tab:teams} lists the $21$ teams. A total of $15$ teams submitted a total of $16$ description papers from which we accepted $15$ papers for publication. Accepted papers are given in Table~\ref{tab:teams}.

\subsection{Baselines}\label{sec:baselines}
We provide three baselines for each of the two subtasks. \textbf{Baseline-I} is based on the majority class in the Train data for each subtask. For Subtask~1, Baseline-I performs at \textit{$F_1$=$1.97$} on Test-A and \textit{$F_1$=$2.59$} on Test-B, hence it obtains an average $F_1$ of $2.28$. For Subtask~2, Baseline-I performs at $F\textsubscript{NP}$=$27.83$. 
\textbf{Baseline-mBERT}, \textbf{Baseline-XLMR}, and \textbf{Baseline-MARBERT} are fine-tuned multi-lingual BERT-Base model (mBERT)~\cite{devlin-2019-bert}, cross-lingual RoBERTa (XLMR)~\cite{crosslingual-2019-conneau}, and MARBERT~\cite{mageed2020marbert}, respectively. More specifically, we take checkpoints for these models from Hugginface Library~\cite{wolf-2020-transformers} and fine-tune each of them for $20$ epochs with a learning rate of $2e$-$5$ and batch size of $32$. The maximum length of input sequence is set to $64$ tokens. We evaluate each model at the end of each epoch and choose the best model based on performance on the respective Dev set. We then report performance of the best model on test sets. Baseline-MARBERT is our strongest baseline: it obtains \textit{$F_1$=$31.39$} on Test-A of Subtask~1, \textit{$F_1$=$16.94$} on Test-B of Subtask~1, average \textit{$F_1$=$24.17$} over Test-A and Test-B, and $F\textsubscript{NP}$=$72.36$ on Subtask~2. 

\input{tables/subtask1_res}

\input{tables/subtask1-A}
\input{tables/subtask1-B}
\input{tables/subtask2_res}

\subsection{Shared Task Results}\label{sec:task_results}
Table~\ref{tab:sub1_res} presents the leaderboard of Subtask~1 and is sorted by the main metric of Subtask~1, i.e., average macro-$F_1$ score. As Tables~\ref{tab:sub1a_res} and~\ref{tab:sub1b_res} show, for each team, we take their best score of Test-A and Test-B and then calculate the average macro-$F_1$ score over the best scores of these two test sets (i.e., Test-A and Test-B). \texttt{Team rematchka}~\cite{abdelsalam2022dialect} obtained the best performance on Subtask~1 with $27.06$ average macro-$F_1$. We can observe that seven teams outperform our strongest baseline, Baseline-MARBERT. \texttt{Team rematchka} also achieved the best $F_1$ of $36.48$ on Test-A of Subtask~1. \texttt{Team UniManc}~\cite{Khered-2022-building} acquired the best $F_1$ of $18.95$ on Test-B of Subtask~1. Results show that dialect identification based on text input is challenging. We note that there is a sizable discrepancy between test results on Test-A and Test-B: Test-B results are much lower. We believe the reason is that Test-B is derived from a different distribution (e.g., different collection time) as compared to training data of Subtask~1.

Table~\ref{tab:sub2_res} shows the leaderboard of Subtask~2 and is sorted by the main metric of Subtask~2, $F_{NP}$ score. Again, \texttt{Team rematchka} achieved the best $F_{NP}$ score of $75.16$. We observe that four and then eight teams outperformed our Baseline-MARBERT and Baseline-XLMR, respectively.

\input{tables/submit_system}

\subsection{General Description of Submitted Systems} 

In Table~\ref{tab:system_sum}, we provide a high-level summary of the submitted systems. For each team, we list their best score with the the main metric of each subtask and the number of their submissions. As shown in this table, most teams used Transformer-based pre-trained language models, including mBERT~\cite{devlin-2019-bert}, ArabBERT~\cite{antoun2020arabert}, MARBERT~\cite{mageed2020marbert}. 

The top team of Subtasks~1 and 2, i.e., \texttt{rematchka}, exploited MARBERT, AraBERT, and AraGPT2~\cite{antoun2021aragpt2} with different prompting techniques and added linguistic features to their models. 

The team placing first on Test-B of Subtask~1, i.e., \texttt{UniManc}, used MARBERT and enhanced the model on under-represented classes by introducing a sampling strategy. 

Teams \texttt{mtu\_fiz}~\cite{shammary-2022-tfidf} and \texttt{ISL\_AAST} used adapter modules to fine-tune MARBERT and applied data augmentation techniques. 

\texttt{Team UniManc} found that further pre-training MARBERT on the $10$M unlabelled tweets we released does not benefit Subtask~1 but improves  performance on Subtask~2. 

Six teams also utilized classical machine learning methods (e.g., SVM and Naive Bayes) to develop their systems.

%% file: tables/subtask1_res.tex
\begin{table}[ht!]
\small
\centering
\setlength{\tabcolsep}{2pt}
\begin{tabular}{rlr}
\toprule
\textbf{}  & \textbf{Team}       & \textbf{Avg. Macro-F\textsubscript{1}} \\ \midrule
1              & rematchka           & 27.06                  \\
2              & UniManc             & 26.86                  \\
3              & GOF                 & 26.44                  \\
4              & mtu\_fiz            & 25.50                  \\
5              & iCompass            & 25.32                  \\
6              & ISL-AAST            & 24.59                  \\
7              & Ahmed\_and\_Khalil  & 24.35                  \\
\multicolumn{2}{l}{\color{blue}Baseline-MARBERT} & 24.17                  \\
8              & Pythoneers          & 24.12                  \\
9              & Giyaseddin          & 22.42                  \\
10             & SQU                 & 22.42                  \\
11             & Elyadata            & 22.41                  \\
12             & NLP\_DI             & 21.28                  \\
13             & RUTeam              & 17.28                  \\
14             & 259                 & 16.89                  \\
15             & zTeam               & 16.12                  \\
16             & XY                  & 15.80                  \\
\multicolumn{2}{l}{\color{blue}Baseline-mBERT}   & 15.70                  \\
17             & BFCAI               & 15.48                  \\
18             & SUKI                & 15.11                  \\
\multicolumn{2}{l}{\color{blue}Baseline-XLMR}    & 14.68                  \\
19             & Oscar\_Garibo       & 14.45                  \\
\multicolumn{2}{l}{\color{blue}Baseline-I}       & 2.28                   \\ \bottomrule
\end{tabular}
\caption{Results for Subtask 1 (Country-Level DA).}\label{tab:sub1_res}
\end{table}

%% file: tables/subtask1-A.tex
\begin{table}[ht!]
\small
\centering
\setlength{\tabcolsep}{3pt}
\begin{tabular}{rlrrrr}
\toprule
\textbf{}  & \textbf{Team}       & \textbf{Macro-F\textsubscript{1}} & \textbf{Acc} & \textbf{Rec} & \textbf{Prec} \\ \midrule
1              & rematchka           & 36.48             & 53.05             & 35.22           & 41.89              \\
2              & GOF                 & 35.68             & 52.10             & 34.91           & 39.18              \\
3              & UniManc             & 34.78             & 52.33             & 34.74           & 38.74              \\
4              & iCompass            & 33.70             & 51.91             & 33.71           & 35.86              \\
5              & mtu\_fiz            & 33.32             & 51.18             & 32.42           & 38.87              \\
6              & Pythoneers          & 32.63             & 48.91             & 31.77           & 36.77              \\
7              & ISL\_AAST           & 32.24             & 50.27             & 32.07           & 37.53              \\
8              & Ahmed\_and\_Khalil  & 31.54             & 50.34             & 32.04           & 34.00              \\
\multicolumn{2}{l}{\color{blue}Baseline-MARBERT} & 31.39             & 47.77             & 31.01           & 35.53              \\
9              & Giyaseddin          & 30.55             & 47.65             & 30.04           & 34.18              \\
10             & SQU                 & 30.01             & 46.85             & 29.75           & 34.57              \\
11             & Elyadata            & 29.35             & 45.84             & 28.60           & 31.27              \\
12             & NLP\_DI             & 26.12             & 42.08             & 25.75           & 28.29              \\
13             & RUTeam              & 23.20             & 36.61             & 22.84           & 24.00              \\
14             & XY                  & 22.36             & 39.85             & 21.33           & 30.52              \\
15             & 259                 & 21.93             & 34.11             & 22.69           & 22.32              \\
16             & zTeam               & 21.76             & 39.43             & 20.77           & 27.25              \\
17             & BFCAI               & 21.25             & 38.63             & 20.47           & 25.25              \\
\multicolumn{2}{l}{\color{blue}Baseline-mBERT}   & 20.88             & 35.22             & 20.67           & 21.82              \\
18             & Oscar\_Garibo       & 20.50             & 36.80             & 20.06           & 22.15              \\
\multicolumn{2}{l}{\color{blue}Baseline-XLMR}    & 19.74             & 36.22             & 19.83           & 21.00              \\
19             & SUKI                & 19.63             & 29.23             & 20.85           & 21.95              \\
\multicolumn{2}{l}{\color{blue}Baseline-I}       & 1.97              & 21.54             & 5.55            & 1.20               \\ \bottomrule
\end{tabular}
\caption{Results on Test-A of Subtask 1.}\label{tab:sub1a_res}
\end{table}

%% file: tables/subtask1-B.tex
\begin{table}[ht!]
\small
\setlength{\tabcolsep}{2pt}

\centering
\begin{tabular}{rlrrrr}
\toprule
\textbf{} & \multicolumn{1}{c}{\textbf{Team}} & \textbf{Macro-F\textsubscript{1}} & \textbf{Acc } & \textbf{Rec } & \textbf{Prec } \\ \midrule
1             & UniManc                           & 18.95             & 36.84             & 20.48           & 25.82              \\
2             & mtu\_fiz                          & 17.67             & 33.92             & 18.79           & 25.03              \\
3             & rematchka                         & 17.64             & 36.50             & 19.62           & 23.59              \\
4             & GOF                               & 17.19             & 34.60             & 18.56           & 22.12              \\
5             & Ahmed\_and\_Khalil                & 17.15             & 34.67             & 19.47           & 23.39              \\
6             & ISL-AAST                          & 16.95             & 35.07             & 18.40           & 22.47              \\
7             & iCompass                          & 16.94             & 34.94             & 19.52           & 19.01              \\
\multicolumn{2}{l}{\color{blue}Baseline-MARBERT} & 16.94             &  34.06           &   18.82          &  23.19            \\
8             & NLP\_DI                           & 16.44             & 27.68             & 18.49           & 20.28              \\
9             & Pythoneers                        & 15.61             & 29.51             & 15.90           & 19.51              \\
10            & Elyadata                          & 15.46             & 29.85             & 16.34           & 20.25              \\
11            & SQU                               & 14.84             & 30.12             & 16.80           & 21.32              \\
12            & Giyaseddin                        & 14.30             & 29.92             & 15.59           & 21.95              \\
13            & 259                               & 11.85             & 22.25             & 11.43           & 14.21              \\
14            & RUTeam                            & 11.35             & 22.80             & 11.86           & 14.60              \\
15            & SUKI                              & 10.58             & 20.56             & 10.11           & 12.98              \\
\multicolumn{2}{l}{\color{blue}Baseline-mBERT}    & 10.53             & 22.05            &  11.42          &  14.06              \\
16            & zTeam                             & 10.47             & 25.71             & 13.23           & 16.29              \\
17            & BFCAI                             & 9.71              & 23.13             & 11.99           & 14.54              \\
\multicolumn{2}{l}{\color{blue}Baseline-XLMR}     & 9.62              & 21.91            & 11.33           & 14.05            \\
18            & XY                                & 9.25              & 23.74             & 11.73           & 17.57              \\
19            & Oscar\_Garibo                     & 8.40              & 19.40             & 9.80            & 11.74              \\
\multicolumn{2}{l}{\color{blue}Baseline-I}       & 2.59           &   14.86                    &   10.00                 &  1.49                      \\ \bottomrule
\end{tabular}
\caption{Results on Test-B of Subtask 1.}\label{tab:sub1b_res}
\end{table}

%% file: tables/subtask2_res.tex
\begin{table}[ht!]
\small
\setlength{\tabcolsep}{3pt}

\centering
\begin{tabular}{rlrrrr}
\toprule
\textbf{} & \multicolumn{1}{c}{\textbf{Team}} & \multicolumn{1}{c}{\textbf{F\textsubscript{1}-PN}} & \multicolumn{1}{c}{\textbf{Acc}} & \multicolumn{1}{c}{\textbf{Rec}} & \multicolumn{1}{c}{\textbf{Prec}} \\ \midrule
1             & rematchka                         & 75.16                                    & 69.70                                 & 66.22                               & 67.57                                  \\
2             & UniManc                           & 73.54                                    & 67.70                                 & 63.92                               & 65.27                                  \\
3             & BhamNLP                           & 73.46                                    & 67.33                                 & 62.83                               & 65.24                                  \\
4             & Pythoneers                        & 73.40                                    & 68.23                                 & 65.87                               & 66.08                                  \\
\multicolumn{2}{l}{\color{blue}Baseline-MARBERT}  & 72.36                                    & 66.66                              &  63.92                           & 64.50                                 \\
5             & Ahmed\_and\_Khalil                & 71.46                                    & 66.03                                 & 63.73                               & 63.84                                  \\
6             & Giyaseddin                        & 71.43                                    & 65.80                                 & 62.20                               & 63.51                                  \\
7             & ISL\_AAST                         & 70.55                                    & 64.97                                 & 61.41                               & 62.58                                  \\
8             & ANLP-RG                           & 67.31                                    & 61.90                                 & 59.67                               & 59.69                                  \\
\multicolumn{2}{l}{\color{blue}Baseline-XLMR}           & 63.24                              & 57.30                                 & 55.53                              &  55.66                                 \\
9             & RUTeam                            & 61.07                                    & 56.17                                 & 53.58                               & 53.90                                  \\
\multicolumn{2}{l}{\color{blue}Baseline-mBERT}          & 55.84                              & 50.13                                & 49.00                             & 49.47                              \\
10            & Oscar\_Garibo                     & 46.43                                    & 43.00                                 & 41.92                               & 42.00                                  \\
\multicolumn{2}{l}{\color{blue}Baseline-I}       & 27.83                                    &  38.57                               &  33.33                            & 12.86                              \\ \bottomrule
\end{tabular}
\caption{Results for Subtask 2 (Sentiment Analysis).}\label{tab:sub2_res}
\end{table}

%% file: tables/submit_system.tex
\begin{table*}[ht!]
\centering
\small
\setlength{\tabcolsep}{0pt}

\begin{tabular}{lcrccccccccccccccc}
\toprule
\multicolumn{1}{c}{\multirow{2}{*}{\textbf{}}} & \multirow{2}{*}{\textbf{\rotatebox[origin=c]{70}{\# submit}}} & \multicolumn{1}{c}{\multirow{2}{*}{\textbf{\rotatebox[origin=c]{70}{Main Metric}}}} & \multicolumn{5}{c}{\textbf{Features}}                                                                    & \multicolumn{9}{c}{\textbf{Techniques}}                                                                                                                                                          & \textbf{Use unlabeled} \\ \cmidrule(l){4-8} \cmidrule(l){9-17} \cmidrule(l){18-18} 
\multicolumn{1}{c}{\multirow{2}{*}{\textbf{Team Name}}}                                    &                                     & \multicolumn{1}{c}{}                                      & \textbf{\rotatebox[origin=c]{70}{$N$-gram}}      & \textbf{\rotatebox[origin=c]{70}{TF-IDF}} & \textbf{\rotatebox[origin=c]{70}{Linguistic}} & \textbf{\rotatebox[origin=c]{70}{Word embeds}} & \textbf{\rotatebox[origin=c]{70}{Sampling}} & \textbf{\rotatebox[origin=c]{70}{Classical ML}} & \textbf{\rotatebox[origin=c]{70}{Neural nets}} & \textbf{\rotatebox[origin=c]{70}{Transformer}} & \textbf{\rotatebox[origin=c]{70}{Ensemble}} & \textbf{\rotatebox[origin=c]{70}{Adapter}} & \textbf{\rotatebox[origin=c]{70}{Multitask}} & \textbf{\rotatebox[origin=c]{70}{Prompting}} & \textbf{\rotatebox[origin=c]{70}{Distillation}} & \textbf{\rotatebox[origin=c]{70}{Data Aug.}} & \textbf{\rotatebox[origin=c]{70}{Pre-training}}         \\ \midrule
\multicolumn{18}{c}{\textbf{\colorbox{yellow!15}{Subtask 1}}}                                                                                                                                                                                                                                                                                                                                                                                                                                                              \\ \midrule
\textbf{rematchka}                                      & 6                                   & 27.06                                                     &                      &                 &                      &                      &                   &                       & \checkmark                    & \checkmark                    & \checkmark                 &                  &                    & \checkmark                  &                       &                   &                            \\
\textbf{UniManc}                                        & 6                                   & 26.86                                                     &                      &                 &                      &                      & \checkmark                 &                       &                      & \checkmark                    &                   &                  &                    &                    &                       &                   &                            \\
\textbf{GOF}                                            & 4                                   & 26.44                                                     &                      &                 &                      &                      &                   &                       &                      & \checkmark                    &                   &                  &                    &                    &                       &                   &                            \\
\textbf{mtu\_fiz}                                       & 8                                   & 25.50                                                     &                      &                 &                      &                      &                   &                       & \checkmark                    & \checkmark                    &                   & \checkmark                &                    &                    &                       & \checkmark                 &                            \\
\textbf{iCompass}                                       & 2                                   & 25.32                                                     &                      &                 &                      &                      &                   &                       &                      & \checkmark                    &                   &                  &                    &                    &                       &                   &                            \\
\textbf{ISL\_AAST}                                      & 5                                   & 24.59                                                     &                      &                 &                      &                      &                   &                       &                      & \checkmark                    &                   & \checkmark                &                    &                    &                       & \checkmark                 &                            \\
\textbf{Ahmed\_and\_Khalil}                             & 2                                   & 24.35                                                     &                      &                 &                      &                      &                   &                       &                      & \checkmark                    &                   &                  &                    &                    &                       &                   &                            \\
\textbf{Pythoneers}                                     & 4                                   & 24.12                                                     &                      &                 &                      &                      &                   &                       &                      & \checkmark                    & \checkmark                 &                  & \checkmark                  &                    & \checkmark                     &                   &                            \\
\textbf{Giyaseddin}                                     & 3                                   & 22.42                                                     &                      &                 &                      &                      &                   &                       &                      & \checkmark                    &                   &                  &                    &                    &                       &                   &                            \\
\textbf{SQU}                                            & 4                                   & 22.42                                                     & \checkmark                    & \checkmark               &                      & \checkmark                    &                   & \checkmark                     & \checkmark                    & \checkmark                    &                   &                  &                    &                    &                       &                   & \checkmark                          \\
\textbf{NLP\_DI}                                        & 9                                   & 21.28                                                     & \checkmark                    &                 &                      &                      &                   &                       & \checkmark                    & \checkmark                    &                   &                  &                    &                    &                       &                   & \checkmark                          \\
\textbf{RUTeam}                                         & 2                                   & 17.28                                                      &                      &                 &       \checkmark               &                      &                   &                       &                    &            \checkmark            &                   &                  &                    &                    &                       &                   &                            \\
\textbf{259}                                            & 2                                   & 16.89                                                      & \checkmark                    &                 &                      &                      &                   & \checkmark                     &                      & \checkmark                    &                   &                  &                    &                    &                       &                   &                            \\
\textbf{zTeam}                                          & 2                                   & 16.12                                                      &                      &                 &                      & \checkmark                    &                   & \checkmark                     & \checkmark                    & \checkmark                    &                   &                  &                    &                    &                       &                   &                            \\
\textbf{XY}                                             & 10                                  & 15.80                                                      &                      &                 &                      &                      &                   & \checkmark                     & \checkmark                    & \checkmark                    & \checkmark                 &                  &                    &                    &                       &                   &                            \\
\textbf{BFCAI}                                          & 6                                   & 15.48                                                      &                      & \checkmark               &                      & \checkmark                    &                   & \checkmark                     & \checkmark                    &                      &                   &                  &                    &                    &                       &                   &                            \\
\textbf{SUKI}                                           & 2                                   & 15.11                                                      & \checkmark                    &                 &                      &                      &                   & \checkmark                     &                      &                      &                   &                  &                    &                    &                       &                   &                            \\

\midrule
\multicolumn{18}{c}{\textbf{\colorbox{green!10}{Subtask 2}}}                                                                                                                                                                                                                                                                                                                                                                                                                                                              \\ \midrule
\textbf{rematchka}                                      & 4                                   & 75.16                                                     &                      &                 & \checkmark                    &                      &                   &                       & \checkmark                    & \checkmark                    & \checkmark                 &                  &                    & \checkmark                  &                       &                   &                            \\
\textbf{UniManc}                                        & 3                                   & 73.54                                                     &                      &                 &                      &                      &                   &                       &                      & \checkmark                    &                   &                  &                    &                    &                       &                   & \checkmark                          \\
\textbf{BhamNLP}                                        & 3                                   & 73.46                                                     & \checkmark                    &                 &                      & \checkmark                    &                   &                       & \checkmark                    & \checkmark                    &                   &                  &                    &                    &                       &                   &                            \\
\textbf{Pythoneers}                                     & 1                                   & 73.40                                                     &                      &                 &                      &                      &                   &                       &                      & \checkmark                    & \checkmark                 &                  & \checkmark                  &                    & \checkmark                     &                   &                            \\
\textbf{Ahmed\_and\_Khalil}                             & 1                                   & 71.46                                                     &                      &                 &                      &                      &                   &                       &                      & \checkmark                    &                   &                  &                    &                    &                       &                   &                            \\
\textbf{Giyaseddin}                                     & 1                                   & 71.43                                                     &                      &                 &                      &                      &                   &                       &                      & \checkmark                    &                   &                  &                    &                    &                       &                   &                            \\
\textbf{ISL\_AAST}                                      & 3                                   & 70.55                                                     &                      &                 &                      &                      &                   &                       &                      & \checkmark                    &                   & \checkmark                &                    &                    &                       & \checkmark                 &                            \\
\textbf{ANLP-RG}                                        & 3                                   & 67.31                                                     &                      &                 &                      &                      &                   &                       &                      & \checkmark                    &                   &                  &                    &                    &                       &                   &                            \\
\textbf{RUTeam}                                         & 1                                   & 61.07                                                      &                      &                 &       \checkmark               &                      &                   &                       &                    &            \checkmark            &                   &                  &                    &                    &                       &                   &                            \\
\bottomrule
\end{tabular}
\caption{Summary of approaches used by participating teams who also submitted system descriptions. Teams are sorted by their performance on official metric, the average $Macro$-$F_1$ score over Test-A and Test-B for Subtask~1 and $F1_{NP}$ score over the positive and negative classes for Subtask~2. Classical machine learning (ML) refers to any non-neural machine learning methods such as naive Bayes and support vector machines. The term ``neural nets" refers to any model based on neural networks (e.g., FFNN, RNN, and CNN) except Transformer models. Transformer refers to neural networks based on a Transformer architecture such as BERT. \textbf{Data Aug.}: Data Augmentation.}
\label{tab:system_sum}
\end{table*}

%% file: conc.tex
\section{Conclusion and Future Work}
\label{sec:conc}

We presented the findings and results of the third Nuanced Arabic Dialect Identification shared task, NADI 2022. The shared task has two subtasks: Subtask 1 on country-level dialect identification (including $18$ countries) and Subtask 2 on dialectal sentiment analysis (including $10$ countries). NADI continues to be an attractive shared task, as reflected by the wide participation: $41$ registered teams, $21$ submitting teams scoring $105$ valid models, and $15$ published papers. Results obtained by the various teams show that both dialect identification and dialectal sentiment analysis of short text sequences remain challenging tasks. This motivates further work on Arabic dialects, and so we plan to run future iterations of NADI. Our experience from NADI 2022 shows that inclusion of additional subtasks, along with dialect identification, provides a rich context for modeling. Hence, we intend to continue adding at least one subtask (e.g., sentiment analysis covering more countries, emotion detection) to our main focus of dialect identification. We will also consider adding a data contribution track to NADI. In that track, teams may collect and label new datasets for public release.

%% file: ack.tex
\section*{Acknowledgements}\label{sec:acknow}
MAM acknowledges support from Canada Research Chairs (CRC), the Natural Sciences and Engineering Research Council of Canada (NSERC; RGPIN-2018-04267), the Social Sciences and Humanities Research Council of Canada (SSHRC; 435-2018-0576; 895-2020-1004; 895-2021-1008), Canadian Foundation for Innovation (CFI; 37771), and Digital Research Alliance of Canada,\footnote{\href{https://alliancecan.ca}{https://alliancecan.ca}} and UBC Advanced Research Computing-Sockeye.\footnote{\href{https://arc.ubc.ca/ubc-arc-sockeye}{https://arc.ubc.ca/ubc-arc-sockeye}}

%% file: main.bbl
\begin{thebibliography}{85}
\expandafter\ifx\csname natexlab\endcsname\relax\def\natexlab#1{#1}\fi

\bibitem[{AAlAbdulsalam(2022)}]{aalabdulsalam-2022-squcs}
Abdulrahman AAlAbdulsalam. 2022.
\newblock {SQU-CS @ NADI 2022: Dialectal Arabic Identification using One-vs-One
  Classification with TF-IDF Weights Computed on Character n-grams}.
\newblock In \emph{Proceedings of the Seventh Arabic Natural Language
  Processing Workshop (WANLP 2022)}. Association for Computational Linguistics.

\bibitem[{Abbes et~al.(2020)Abbes, Zaghouani, El-Hardlo, and
  Ashour}]{abbes2020daict}
Ines Abbes, Wajdi Zaghouani, Omaima El-Hardlo, and Faten Ashour. 2020.
\newblock \href {https://aclanthology.org/2020.lrec-1.768} {{DAICT}: A
  dialectal {A}rabic irony corpus extracted from {T}witter}.
\newblock In \emph{Proceedings of the 12th Language Resources and Evaluation
  Conference}, pages 6265--6271, Marseille, France. European Language Resources
  Association.

\bibitem[{Abdel-Salam(2022)}]{abdelsalam2022dialect}
Reem Abdel-Salam. 2022.
\newblock Dialect \& sentiment identification in nuanced {A}rabic tweets using
  an ensemble of prompt-based, fine-tuned and multitask bert-based models.
\newblock In \emph{Proceedings of the Seventh Arabic Natural Language
  Processing Workshop (WANLP 2022)}. Association for Computational Linguistics.

\bibitem[{Abdelali et~al.(2021)Abdelali, Mubarak, Samih, Hassan, and
  Darwish}]{abdelali-etal-2021-qadi}
Ahmed Abdelali, Hamdy Mubarak, Younes Samih, Sabit Hassan, and Kareem Darwish.
  2021.
\newblock \href {https://aclanthology.org/2021.wanlp-1.1} {{QADI}: {A}rabic
  dialect identification in the wild}.
\newblock In \emph{Proceedings of the Sixth Arabic Natural Language Processing
  Workshop}, pages 1--10, Kyiv, Ukraine (Virtual). Association for
  Computational Linguistics.

\bibitem[{Abdul-Mageed(2015)}]{mageed2015Dissert}
Muhammad Abdul-Mageed. 2015.
\newblock \href
  {https://search.proquest.com/openview/11a055088001fdb6522d6a4094550ad5/1?pq-origsite=gscholar&cbl=18750}
  {\emph{Subjectivity and sentiment analysis of Arabic as a
  morophologically-rich language}}.
\newblock Ph.D. thesis, Indiana University.

\bibitem[{Abdul{-}Mageed(2019)}]{mageed2019modeling}
Muhammad Abdul{-}Mageed. 2019.
\newblock \href {https://doi.org/10.1016/j.ipm.2017.07.004} {Modeling {A}rabic
  subjectivity and sentiment in lexical space}.
\newblock \emph{Information Processing \& Management}, 56(2):291--307.

\bibitem[{Abdul-Mageed et~al.(2018)Abdul-Mageed, Alhuzali, and
  Elaraby}]{Abdul-Mageed:2018:you}
Muhammad Abdul-Mageed, Hassan Alhuzali, and Mohamed Elaraby. 2018.
\newblock \href {https://aclanthology.org/L18-1577} {You tweet what you speak:
  A city-level dataset of {A}rabic dialects}.
\newblock In \emph{Proceedings of the Eleventh International Conference on
  Language Resources and Evaluation ({LREC} 2018)}, Miyazaki, Japan. European
  Language Resources Association (ELRA).

\bibitem[{Abdul{-}Mageed et~al.(2014)Abdul{-}Mageed, Diab, and
  K{\"{u}}bler}]{Abdul-Mageed:2014:samar}
Muhammad Abdul{-}Mageed, Mona~T. Diab, and Sandra K{\"{u}}bler. 2014.
\newblock \href {https://doi.org/10.1016/j.csl.2013.03.001} {{SAMAR:}
  subjectivity and sentiment analysis for arabic social media}.
\newblock \emph{Comput. Speech Lang.}, 28(1):20--37.

\bibitem[{Abdul-Mageed et~al.(2021{\natexlab{a}})Abdul-Mageed, Elmadany, and
  Nagoudi}]{mageed2020marbert}
Muhammad Abdul-Mageed, AbdelRahim Elmadany, and El~Moatez~Billah Nagoudi.
  2021{\natexlab{a}}.
\newblock \href {https://doi.org/10.18653/v1/2021.acl-long.551} {{ARBERT} {\&}
  {MARBERT}: Deep bidirectional transformers for {A}rabic}.
\newblock In \emph{Proceedings of the 59th Annual Meeting of the Association
  for Computational Linguistics and the 11th International Joint Conference on
  Natural Language Processing (Volume 1: Long Papers)}, pages 7088--7105,
  Online. Association for Computational Linguistics.

\bibitem[{Abdul-Mageed et~al.(2020{\natexlab{a}})Abdul-Mageed, Zhang, Bouamor,
  and Habash}]{mageed:2020:nadi}
Muhammad Abdul-Mageed, Chiyu Zhang, Houda Bouamor, and Nizar Habash.
  2020{\natexlab{a}}.
\newblock \href {https://aclanthology.org/2020.wanlp-1.9} {{NADI} 2020: The
  first nuanced {A}rabic dialect identification shared task}.
\newblock In \emph{Proceedings of the Fifth Arabic Natural Language Processing
  Workshop}, pages 97--110, Barcelona, Spain (Online). Association for
  Computational Linguistics.

\bibitem[{Abdul-Mageed et~al.(2021{\natexlab{b}})Abdul-Mageed, Zhang, Elmadany,
  Bouamor, and Habash}]{abdul-mageed-etal-2021-nadi}
Muhammad Abdul-Mageed, Chiyu Zhang, AbdelRahim Elmadany, Houda Bouamor, and
  Nizar Habash. 2021{\natexlab{b}}.
\newblock \href {https://aclanthology.org/2021.wanlp-1.28} {{NADI} 2021: The
  second nuanced {A}rabic dialect identification shared task}.
\newblock In \emph{Proceedings of the Sixth Arabic Natural Language Processing
  Workshop}, pages 244--259, Kyiv, Ukraine (Virtual). Association for
  Computational Linguistics.

\bibitem[{Abdul-Mageed et~al.(2020{\natexlab{b}})Abdul-Mageed, Zhang, Elmadany,
  and Ungar}]{mageed2020microdialect}
Muhammad Abdul-Mageed, Chiyu Zhang, AbdelRahim Elmadany, and Lyle Ungar.
  2020{\natexlab{b}}.
\newblock \href {https://doi.org/10.18653/v1/2020.emnlp-main.472} {Toward
  micro-dialect identification in diaglossic and code-switched environments}.
\newblock In \emph{Proceedings of the 2020 Conference on Empirical Methods in
  Natural Language Processing (EMNLP)}, pages 5855--5876, Online. Association
  for Computational Linguistics.

\bibitem[{Abdul-Mageed et~al.(2020{\natexlab{c}})Abdul-Mageed, Zhang, Hashemi,
  and Nagoudi}]{mageed2020aranet}
Muhammad Abdul-Mageed, Chiyu Zhang, Azadeh Hashemi, and El~Moatez~Billah
  Nagoudi. 2020{\natexlab{c}}.
\newblock \href {https://aclanthology.org/2020.osact-1.3} {{A}ra{N}et: A deep
  learning toolkit for {A}rabic social media}.
\newblock In \emph{Proceedings of the 4th Workshop on Open-Source Arabic
  Corpora and Processing Tools, with a Shared Task on Offensive Language
  Detection}, pages 16--23, Marseille, France. European Language Resource
  Association.

\bibitem[{Abo et~al.(2021)Abo, Idris, Mahmud, Qazi, Hashem, Maitama, Naseem,
  Khan, and Yang}]{abo2021multi}
Mohamed Elhag~Mohamed Abo, Norisma Idris, Rohana Mahmud, Atika Qazi, Ibrahim
  Abaker~Targio Hashem, Jaafar~Zubairu Maitama, Usman Naseem, Shah~Khalid Khan,
  and Shuiqing Yang. 2021.
\newblock \href {https://www.mdpi.com/1261280} {A multi-criteria approach for
  {A}rabic dialect sentiment analysis for online reviews: Exploiting optimal
  machine learning algorithm selection}.
\newblock \emph{Sustainability}, 13(18):10018.

\bibitem[{Abu~Farha and Magdy(2020)}]{farha2020arabic}
Ibrahim Abu~Farha and Walid Magdy. 2020.
\newblock \href {https://aclanthology.org/2020.osact-1.5} {From {A}rabic
  sentiment analysis to sarcasm detection: The {A}r{S}arcasm dataset}.
\newblock In \emph{Proceedings of the 4th Workshop on Open-Source Arabic
  Corpora and Processing Tools, with a Shared Task on Offensive Language
  Detection}, pages 32--39, Marseille, France. European Language Resource
  Association.

\bibitem[{Abu~Farha and Magdy(2021)}]{farha2021benchmarking}
Ibrahim Abu~Farha and Walid Magdy. 2021.
\newblock \href {https://aclanthology.org/2021.wanlp-1.3} {Benchmarking
  transformer-based language models for {A}rabic sentiment and sarcasm
  detection}.
\newblock In \emph{Proceedings of the Sixth Arabic Natural Language Processing
  Workshop}, pages 21--31, Kyiv, Ukraine (Virtual). Association for
  Computational Linguistics.

\bibitem[{Al-Laith et~al.(2021)Al-Laith, Shahbaz, Alaskar, and
  Rehmat}]{al2021arasencorpus}
Ali Al-Laith, Muhammad Shahbaz, Hind~F Alaskar, and Asim Rehmat. 2021.
\newblock \href {https://www.mdpi.com/1027412} {Arasencorpus: A semi-supervised
  approach for sentiment annotation of a large {A}rabic text corpus}.
\newblock \emph{Applied Sciences}, 11(5):2434.

\bibitem[{Al-Sabbagh and Girju(2012)}]{al2012yadac}
Rania Al-Sabbagh and Roxana Girju. 2012.
\newblock \href
  {http://www.lrec-conf.org/proceedings/lrec2012/pdf/663_Paper.pdf} {{YADAC}:
  Yet another dialectal {A}rabic corpus}.
\newblock In \emph{Proceedings of the Eighth International Conference on
  Language Resources and Evaluation ({LREC}'12)}, pages 2882--2889, Istanbul,
  Turkey. European Language Resources Association (ELRA).

\bibitem[{Al{-}Twairesh et~al.(2018)Al{-}Twairesh, Al{-}Matham, Madi, Almugren,
  Al{-}Aljmi, Alshalan, Alshalan, Alrumayyan, Al{-}Manea, Bawazeer,
  Al{-}Mutlaq, Almanea, Huwaymil, Alqusair, Alotaibi, Al{-}Senaydi, and
  Alfutamani}]{Al-Twairesh:2018:suar}
Nora Al{-}Twairesh, Rawan~N. Al{-}Matham, Nora Madi, Nada Almugren, Al{-}Hanouf
  Al{-}Aljmi, Shahad Alshalan, Raghad Alshalan, Nafla Alrumayyan, Shams
  Al{-}Manea, Sumayah Bawazeer, Nourah Al{-}Mutlaq, Nada Almanea, Waad~Bin
  Huwaymil, Dalal Alqusair, Reem Alotaibi, Suha Al{-}Senaydi, and Abeer
  Alfutamani. 2018.
\newblock \href {https://doi.org/10.1016/j.procs.2018.10.462} {{SUAR:} towards
  building a corpus for the saudi dialect}.
\newblock In \emph{Fourth International Conference On Arabic Computational
  Linguistics, {ACLING} 2018, November 17-19, 2018, Dubai, United Arab
  Emirates}, volume 142 of \emph{Procedia Computer Science}, pages 72--82.
  Elsevier.

\bibitem[{Alhumoud and Wazrah(2022)}]{alhumoud2021arabic}
Sarah~Omar Alhumoud and Asma Ali~Al Wazrah. 2022.
\newblock \href {https://doi.org/10.1007/s10462-021-09989-9} {Arabic sentiment
  analysis using recurrent neural networks: a review}.
\newblock \emph{Artif. Intell. Rev.}, 55(1):707--748.

\bibitem[{Alhuzali et~al.(2018)Alhuzali, Abdul-Mageed, and
  Ungar}]{alhuzali2018enabling}
Hassan Alhuzali, Muhammad Abdul-Mageed, and Lyle Ungar. 2018.
\newblock \href {https://doi.org/10.18653/v1/W18-1104} {Enabling deep learning
  of emotion with first-person seed expressions}.
\newblock In \emph{Proceedings of the Second Workshop on Computational Modeling
  of People{'}s Opinions, Personality, and Emotions in Social Media}, pages
  25--35, New Orleans, Louisiana, USA. Association for Computational
  Linguistics.

\bibitem[{Almuqren and Cristea(2021)}]{almuqren2021aracust}
Latifah Almuqren and Alexandra~I. Cristea. 2021.
\newblock \href {https://doi.org/10.7717/peerj-cs.510} {Aracust: a saudi
  telecom tweets corpus for sentiment analysis}.
\newblock \emph{PeerJ Comput. Sci.}, 7:e510.

\bibitem[{Alowisheq et~al.(2021)Alowisheq, Al{-}Twairesh, Altuwaijri,
  AlMoammar, Alsuwailem, Albuhairi, Alahaideb, and
  Alhumoud}]{alowisheq2021marsa}
Areeb Alowisheq, Nora Al{-}Twairesh, Mawaheb Altuwaijri, Afnan AlMoammar,
  Alhanouf Alsuwailem, Tarfa Albuhairi, Wejdan Alahaideb, and Sarah Alhumoud.
  2021.
\newblock \href {https://doi.org/10.1109/ACCESS.2021.3120746} {{MARSA:}
  multi-domain {A}rabic resources for sentiment analysis}.
\newblock \emph{{IEEE} Access}, 9:142718--142728.

\bibitem[{AlShenaifi and Azmi(2022)}]{alshenaifi-2022-arabic}
Nouf AlShenaifi and Aqil Azmi. 2022.
\newblock {Arabic dialect identification using machine learning and
  transformer-based models: Submission to the NADI 2022 Shared Task}.
\newblock In \emph{Proceedings of the Seventh Arabic Natural Language
  Processing Workshop (WANLP 2022)}. Association for Computational Linguistics.

\bibitem[{Alsudais et~al.(2022)Alsudais, Alotaibi, and
  Alomary}]{alsudais2022similarities}
Abdulkareem Alsudais, Wafa Alotaibi, and Faye Alomary. 2022.
\newblock \href {https://arxiv.org/pdf/2105.04221} {Similarities between
  {A}rabic dialects: Investigating geographical proximity}.
\newblock \emph{Information Processing \& Management}, 59(1):102770.

\bibitem[{Althobaiti(2020)}]{althobaiti2020automatic}
Maha~J Althobaiti. 2020.
\newblock \href {https://arxiv.org/abs/2009.12622} {Automatic {A}rabic dialect
  identification systems for written texts: {A} survey}.
\newblock \emph{arXiv preprint arXiv:2009.12622}.

\bibitem[{Althobaiti(2022)}]{althobaiti2022creation}
Maha~J. Althobaiti. 2022.
\newblock \href {https://doi.org/10.1017/S135132492100019X} {Creation of
  annotated country-level dialectal {A}rabic resources: An unsupervised
  approach}.
\newblock \emph{Nat. Lang. Eng.}, 28(5):607--648.

\bibitem[{Alwakid et~al.(2022)Alwakid, Osman, Haj, Alanazi, Humayun, and
  Sama}]{alwakid2022muldasa}
Ghadah Alwakid, Taha Osman, Mahmoud~El Haj, Saad Alanazi, Mamoona Humayun, and
  Najm~Us Sama. 2022.
\newblock \href {https://www.mdpi.com/article/10.3390/app12083806} {{MULDASA}:
  Multifactor lexical sentiment analysis of social-media content in nonstandard
  {A}rabic social media}.
\newblock \emph{Applied Sciences}, 12(8):3806.

\bibitem[{Antoun et~al.(2020)Antoun, Baly, and Hajj}]{antoun2020arabert}
Wissam Antoun, Fady Baly, and Hazem Hajj. 2020.
\newblock \href {https://aclanthology.org/2020.osact-1.2} {{A}ra{BERT}:
  Transformer-based model for {A}rabic language understanding}.
\newblock In \emph{Proceedings of the 4th Workshop on Open-Source Arabic
  Corpora and Processing Tools, with a Shared Task on Offensive Language
  Detection}, pages 9--15, Marseille, France. European Language Resource
  Association.

\bibitem[{Antoun et~al.(2021)Antoun, Baly, and Hajj}]{antoun2021aragpt2}
Wissam Antoun, Fady Baly, and Hazem Hajj. 2021.
\newblock \href {https://aclanthology.org/2021.wanlp-1.21} {{A}ra{GPT}2:
  Pre-trained transformer for {A}rabic language generation}.
\newblock In \emph{Proceedings of the Sixth Arabic Natural Language Processing
  Workshop}, pages 196--207, Kyiv, Ukraine (Virtual). Association for
  Computational Linguistics.

\bibitem[{Attieh and Hassan(2022)}]{attieh-2022-arabic}
Joseph Attieh and Fadi Abdulfattah~Mohammed Hassan. 2022.
\newblock {Arabic Dialect Identification and Sentiment Classification using
  Transformer-based Models}.
\newblock In \emph{Proceedings of the Seventh Arabic Natural Language
  Processing Workshop (WANLP 2022)}. Association for Computational Linguistics.

\bibitem[{Badawi(1973)}]{badawi1973levels}
MS~Badawi. 1973.
\newblock {Levels of contemporary Arabic in Egypt}.
\newblock \emph{Cairo: D{\^a}r al Ma’{\^a}rif}.

\bibitem[{Baimukan et~al.(2022)Baimukan, Bouamor, and
  Habash}]{baimukan2022hierarchical}
Nurpeiis Baimukan, Houda Bouamor, and Nizar Habash. 2022.
\newblock \href {https://aclanthology.org/2022.lrec-1.489} {Hierarchical
  aggregation of dialectal data for {A}rabic dialect identification}.
\newblock In \emph{Proceedings of the Thirteenth Language Resources and
  Evaluation Conference, {LREC} 2022, Marseille, France, 20-25 June 2022},
  pages 4586--4596. European Language Resources Association.

\bibitem[{Bayrak and Issifu(2022)}]{bayrak-2022-domain}
Gıyaseddin Bayrak and Abdul~Majeed Issifu. 2022.
\newblock {Domain-Adapted BERT-based models for Nuanced Arabic Dialect
  Identification and Tweet Sentiment Analysis}.
\newblock In \emph{Proceedings of the Seventh Arabic Natural Language
  Processing Workshop (WANLP 2022)}. Association for Computational Linguistics.

\bibitem[{Bouamor et~al.(2014)Bouamor, Habash, and
  Oflazer}]{Bouamor:2014:multidialectal}
Houda Bouamor, Nizar Habash, and Kemal Oflazer. 2014.
\newblock \href
  {http://www.lrec-conf.org/proceedings/lrec2014/pdf/523_Paper.pdf} {A
  multidialectal parallel corpus of {A}rabic}.
\newblock In \emph{Proceedings of the Ninth International Conference on
  Language Resources and Evaluation ({LREC}'14)}, pages 1240--1245, Reykjavik,
  Iceland. European Language Resources Association (ELRA).

\bibitem[{Bouamor et~al.(2018)Bouamor, Habash, Salameh, Zaghouani, Rambow,
  Abdulrahim, Obeid, Khalifa, Eryani, Erdmann, and
  Oflazer}]{Bouamor:2018:madar}
Houda Bouamor, Nizar Habash, Mohammad Salameh, Wajdi Zaghouani, Owen Rambow,
  Dana Abdulrahim, Ossama Obeid, Salam Khalifa, Fadhl Eryani, Alexander
  Erdmann, and Kemal Oflazer. 2018.
\newblock \href {https://aclanthology.org/L18-1535} {The {MADAR} {A}rabic
  dialect corpus and lexicon}.
\newblock In \emph{Proceedings of the Eleventh International Conference on
  Language Resources and Evaluation ({LREC} 2018)}, Miyazaki, Japan. European
  Language Resources Association (ELRA).

\bibitem[{Brustad(2000)}]{Brustad:2000:syntax}
Kristen Brustad. 2000.
\newblock \emph{The Syntax of Spoken {A}rabic: A Comparative Study of Moroccan,
  Egyptian, Syrian, and Kuwaiti Dialects}.
\newblock Georgetown University Press.

\bibitem[{Conneau and Lample(2019)}]{crosslingual-2019-conneau}
Alexis Conneau and Guillaume Lample. 2019.
\newblock \href
  {https://proceedings.neurips.cc/paper/2019/hash/c04c19c2c2474dbf5f7ac4372c5b9af1-Abstract.html}
  {Cross-lingual language model pretraining}.
\newblock In \emph{Advances in Neural Information Processing Systems 32: Annual
  Conference on Neural Information Processing Systems 2019, NeurIPS 2019,
  December 8-14, 2019, Vancouver, BC, Canada}, pages 7057--7067.

\bibitem[{Cowell(1964)}]{Cowell:1964:reference}
Mark~W. Cowell. 1964.
\newblock \emph{{A Reference Grammar of Syrian {A}rabic}}.
\newblock Georgetown University Press, Washington, D.C.

\bibitem[{Devlin et~al.(2019)Devlin, Chang, Lee, and
  Toutanova}]{devlin-2019-bert}
Jacob Devlin, Ming-Wei Chang, Kenton Lee, and Kristina Toutanova. 2019.
\newblock \href {https://doi.org/10.18653/v1/N19-1423} {{BERT}: Pre-training of
  deep bidirectional transformers for language understanding}.
\newblock In \emph{Proceedings of the 2019 Conference of the North {A}merican
  Chapter of the Association for Computational Linguistics: Human Language
  Technologies, Volume 1 (Long and Short Papers)}, pages 4171--4186,
  Minneapolis, Minnesota. Association for Computational Linguistics.

\bibitem[{Diab et~al.(2010)Diab, Habash, Rambow, Altantawy, and
  Benajiba}]{diab2010colaba}
Mona Diab, Nizar Habash, Owen Rambow, Mohamed Altantawy, and Yassine Benajiba.
  2010.
\newblock \href
  {https://citeseerx.ist.psu.edu/viewdoc/download?doi=10.1.1.705.1024&rep=rep1&type=pdf}
  {{COLABA: Arabic dialect annotation and processing}}.
\newblock In \emph{{LREC workshop on Semitic language processing}}, pages
  66--74.

\bibitem[{El-Haj(2020)}]{el-haj-2020-habibi}
Mahmoud El-Haj. 2020.
\newblock \href {https://aclanthology.org/2020.lrec-1.165} {Habibi - a multi
  dialect multi national {A}rabic song lyrics corpus}.
\newblock In \emph{Proceedings of the 12th Language Resources and Evaluation
  Conference}, pages 1318--1326, Marseille, France. European Language Resources
  Association.

\bibitem[{El-Shangiti and Mrini(2022)}]{elshangiti-2022-ahmed}
Ahmed~Oumar El-Shangiti and Khalil Mrini. 2022.
\newblock {Ahmed and Khalil at NADI 2022: Transfer Learning and Addressing
  Class Imbalance for Arabic Dialect Identification and Sentiment Analysis}.
\newblock In \emph{Proceedings of the Seventh Arabic Natural Language
  Processing Workshop (WANLP 2022)}. Association for Computational Linguistics.

\bibitem[{Elfardy et~al.(2014)Elfardy, Al-Badrashiny, and
  Diab}]{Elfardy:2014:aida}
Heba Elfardy, Mohamed Al-Badrashiny, and Mona Diab. 2014.
\newblock \href {https://doi.org/10.3115/v1/W14-3911} {{AIDA}: Identifying code
  switching in informal {A}rabic text}.
\newblock In \emph{Proceedings of the First Workshop on Computational
  Approaches to Code Switching}, pages 94--101, Doha, Qatar. Association for
  Computational Linguistics.

\bibitem[{Elmadany et~al.(2020)Elmadany, Zhang, Abdul-Mageed, and
  Hashemi}]{elmadany2020leveraging}
AbdelRahim Elmadany, Chiyu Zhang, Muhammad Abdul-Mageed, and Azadeh Hashemi.
  2020.
\newblock \href {https://aclanthology.org/2020.osact-1.17.pdf} {Leveraging
  affective bidirectional transformers for offensive language detection}.
\newblock In \emph{Proceedings of the 4th Workshop on Open-Source Arabic
  Corpora and Processing Tools, with a Shared Task on Offensive Language
  Detection}, pages 102--108.

\bibitem[{Elnagar et~al.(2021)Elnagar, Yagi, Nassif, Shahin, and
  Salloum}]{elnagar2021sentiment}
Ashraf Elnagar, Sane Yagi, Ali~Bou Nassif, Ismail Shahin, and Said~A. Salloum.
  2021.
\newblock \href {https://doi.org/10.1007/978-3-030-69717-4\_39} {Sentiment
  analysis in dialectal {A}rabic: {A} systematic review}.
\newblock In \emph{Advanced Machine Learning Technologies and Applications -
  Proceedings of {AMLTA} 2021, Cairo, Egypt, March 22-24, 2021}, volume 1339 of
  \emph{Advances in Intelligent Systems and Computing}, pages 407--417.
  Springer.

\bibitem[{Fourati et~al.(2020)Fourati, Messaoudi, and
  Haddad}]{fourati2020tunizi}
Chayma Fourati, Abir Messaoudi, and Hatem Haddad. 2020.
\newblock \href {https://arxiv.org/abs/2004.14303} {Tunizi: a tunisian arabizi
  sentiment analysis dataset}.
\newblock \emph{arXiv preprint arXiv:2004.14303}.

\bibitem[{Fsih et~al.(2022)Fsih, Kchaou, Boujelbane, and
  Belguith}]{fsih-2022-benchmarking}
Emna Fsih, Sam\'eh Kchaou, Rahma Boujelbane, and Lamia~Hadrich Belguith. 2022.
\newblock {Benchmarking Transfer Learning Approaches for Sentiment Analysis of
  Arabic Dialect}.
\newblock In \emph{Proceedings of the Seventh Arabic Natural Language
  Processing Workshop (WANLP 2022)}. Association for Computational Linguistics.

\bibitem[{Gadalla et~al.(1997)Gadalla, Kilany, Arram, Yacoub, El-Habashi,
  Shalaby, Karins, Rowson, MacIntyre, Kingsbury, Graff, and
  McLemore}]{Gadalla:1997:callhome}
Hassan Gadalla, Hanaa Kilany, Howaida Arram, Ashraf Yacoub, Alaa El-Habashi,
  Amr Shalaby, Krisjanis Karins, Everett Rowson, Robert MacIntyre, Paul
  Kingsbury, David Graff, and Cynthia McLemore. 1997.
\newblock \href {https://catalog.ldc.upenn.edu/LDC97T19} {{CALLHOME} {E}gyptian
  {A}rabic transcripts {LDC97T19}}.
\newblock Web Download. Philadelphia: Linguistic Data Consortium.

\bibitem[{Guellil et~al.(2021)Guellil, Adeel, Azouaou, Benali, Hachani,
  Dashtipour, Gogate, Ieracitano, Kashani, and Hussain}]{guellil2021semi}
Imane Guellil, Ahsan Adeel, Fai{\c{c}}al Azouaou, Fodil Benali, Ala{-}Eddine
  Hachani, Kia Dashtipour, Mandar Gogate, Cosimo Ieracitano, Reza Kashani, and
  Amir Hussain. 2021.
\newblock \href {https://doi.org/10.1007/s42979-021-00510-1} {A semi-supervised
  approach for sentiment analysis of arab(ic+izi) messages: Application to the
  algerian dialect}.
\newblock \emph{{SN} Comput. Sci.}, 2(2):118.

\bibitem[{Guellil et~al.(2020{\natexlab{a}})Guellil, Azouaou, and
  Chiclana}]{guellil2020arautosenti}
Imane Guellil, Faical Azouaou, and Francisco Chiclana. 2020{\natexlab{a}}.
\newblock \href {https://link.springer.com/article/10.1007/s13278-020-00688-x}
  {Arautosenti: Automatic annotation and new tendencies for sentiment
  classification of arabic messages}.
\newblock \emph{Social Network Analysis and Mining}, 10(1):1--20.

\bibitem[{Guellil et~al.(2020{\natexlab{b}})Guellil, Mendoza, and
  Azouaou}]{guellil2020arabic}
Imane Guellil, Marcelo Mendoza, and Fai{\c{c}}al Azouaou. 2020{\natexlab{b}}.
\newblock \href {https://doi.org/10.4114/intartif.vol23iss65pp124-135} {Arabic
  dialect sentiment analysis with {ZERO} effort.
  {\textbackslash}{\textbackslash} case study: Algerian dialect}.
\newblock \emph{Inteligencia Artif.}, 23(65):124--135.

\bibitem[{Habash et~al.(2021)Habash, Bouamor, Hajj, Magdy, Zaghouani, Bougares,
  Tomeh, Abu~Farha, and Touileb}]{wanlp-2021-arabic}
Nizar Habash, Houda Bouamor, Hazem Hajj, Walid Magdy, Wajdi Zaghouani, Fethi
  Bougares, Nadi Tomeh, Ibrahim Abu~Farha, and Samia Touileb, editors. 2021.
\newblock \href {https://aclanthology.org/2021.wanlp-1.0} {\emph{Proceedings of
  the Sixth Arabic Natural Language Processing Workshop}}. Association for
  Computational Linguistics, Kyiv, Ukraine (Virtual).

\bibitem[{Habash(2010)}]{Habash:2010:introduction}
Nizar~Y Habash. 2010.
\newblock \emph{Introduction to {A}rabic natural language processing},
  volume~3.
\newblock Morgan \& Claypool Publishers.

\bibitem[{Harrat et~al.(2014)Harrat, Meftouh, Abbas, and
  Sma{\"{\i}}li}]{Smaili:2014:building}
Salima Harrat, Karima Meftouh, Mourad Abbas, and Kamel Sma{\"{\i}}li. 2014.
\newblock \href
  {http://www.isca-speech.org/archive/interspeech\_2014/i14\_2123.html}
  {Building resources for algerian arabic dialects}.
\newblock In \emph{{INTERSPEECH} 2014, 15th Annual Conference of the
  International Speech Communication Association, Singapore, September 14-18,
  2014}, pages 2123--2127. {ISCA}.

\bibitem[{Harrell(1962)}]{Harrell:1962:short}
R.S. Harrell. 1962.
\newblock \emph{A Short Reference Grammar of Moroccan {A}rabic: With Audio CD}.
\newblock Georgetown classics in {A}rabic language and linguistics. Georgetown
  University Press.

\bibitem[{Hassan et~al.(2021)Hassan, Mubarak, Abdelali, and
  Darwish}]{hassan2021asad}
Sabit Hassan, Hamdy Mubarak, Ahmed Abdelali, and Kareem Darwish. 2021.
\newblock \href {https://aclanthology.org/2021.eacl-demos.14} {{ASAD}: {A}rabic
  social media analytics and un{D}erstanding}.
\newblock In \emph{Proceedings of the 16th Conference of the European Chapter
  of the Association for Computational Linguistics: System Demonstrations},
  pages 113--118, Online. Association for Computational Linguistics.

\bibitem[{Holes(2004)}]{Holes:2004:modern}
Clive Holes. 2004.
\newblock \emph{Modern {A}rabic: Structures, Functions, and Varieties}.
\newblock Georgetown Classics in {A}rabic Language and Linguistics. Georgetown
  University Press.

\bibitem[{Issa et~al.(2021)Issa, AlShakhori1, Al-Bahrani, and
  Hahn-Powell}]{issa-etal-2021-country}
Elsayed Issa, Mohammed AlShakhori1, Reda Al-Bahrani, and Gus Hahn-Powell. 2021.
\newblock \href {https://aclanthology.org/2021.wanlp-1.32} {Country-level
  {A}rabic dialect identification using {RNN}s with and without linguistic
  features}.
\newblock In \emph{Proceedings of the Sixth Arabic Natural Language Processing
  Workshop}, pages 276--281, Kyiv, Ukraine (Virtual). Association for
  Computational Linguistics.

\bibitem[{Jamal et~al.(2022)Jamal, Kassem, Mohamed, and
  Ashraf}]{jamal-2022-arabic}
Salma Jamal, Aly~M. Kassem, Omar Mohamed, and Ali Ashraf. 2022.
\newblock {On The Arabic Dialect}.
\newblock In \emph{Proceedings of the Seventh Arabic Natural Language
  Processing Workshop (WANLP 2022)}. Association for Computational Linguistics.

\bibitem[{Jarrar et~al.(2016)Jarrar, Habash, Alrimawi, Akra, and
  Zalmout}]{Jarrar:2016:curras}
Mustafa Jarrar, Nizar Habash, Faeq Alrimawi, Diyam Akra, and Nasser Zalmout.
  2016.
\newblock \href {https://link.springer.com/article/10.1007/s10579-016-9370-7}
  {{Curras: an annotated corpus for the Palestinian {A}rabic dialect}}.
\newblock \emph{Language Resources and Evaluation}, pages 1--31.

\bibitem[{Jauhiainen et~al.(2022)Jauhiainen, Jauhiainen, and
  Lind\'en}]{jauhiainen-2022-optimizing}
Tommi Jauhiainen, Heidi Jauhiainen, and Krister Lind\'en. 2022.
\newblock {Optimizing Naive Bayes for Arabic Dialect Identification}.
\newblock In \emph{Proceedings of the Seventh Arabic Natural Language
  Processing Workshop (WANLP 2022)}. Association for Computational Linguistics.

\bibitem[{Kanjirangat et~al.(2022)Kanjirangat, Samardzic, Dolamic, and
  Rinaldi}]{kanjirangat-2022-nlpdi}
Vani Kanjirangat, Tanja Samardzic, Ljiljana Dolamic, and Fabio Rinaldi. 2022.
\newblock {NLP\_DI at NADI Shared Task Subtask-1: Sub-word Level Convolutional
  Neural Models and Pre-trained Binary Classifiers for Dialect Identification}.
\newblock In \emph{Proceedings of the Seventh Arabic Natural Language
  Processing Workshop (WANLP 2022)}. Association for Computational Linguistics.

\bibitem[{Khalifa et~al.(2016)Khalifa, Habash, Abdulrahim, and
  Hassan}]{Khalifa:2016:large}
Salam Khalifa, Nizar Habash, Dana Abdulrahim, and Sara Hassan. 2016.
\newblock \href {https://aclanthology.org/L16-1679} {A large scale corpus of
  {G}ulf {A}rabic}.
\newblock In \emph{Proceedings of the Tenth International Conference on
  Language Resources and Evaluation ({LREC}'16)}, pages 4282--4289,
  Portoro{\v{z}}, Slovenia. European Language Resources Association (ELRA).

\bibitem[{Khered et~al.(2022)Khered, Abdelhalim, and
  Batista-Navarro}]{Khered-2022-building}
Abdullah Khered, Ingy Abdelhalim, and Riza Batista-Navarro. 2022.
\newblock {Building an Ensemble of Transformer Models for Arabic Dialect
  Classification and Sentiment Analysis}.
\newblock In \emph{Proceedings of the Seventh Arabic Natural Language
  Processing Workshop (WANLP 2022)}. Association for Computational Linguistics.

\bibitem[{Malmasi et~al.(2016)Malmasi, Zampieri, Ljube{\v{s}}i{\'c}, Nakov,
  Ali, and Tiedemann}]{malmasi2016discriminating}
Shervin Malmasi, Marcos Zampieri, Nikola Ljube{\v{s}}i{\'c}, Preslav Nakov,
  Ahmed Ali, and J{\"o}rg Tiedemann. 2016.
\newblock \href {https://aclanthology.org/W16-4801} {Discriminating between
  similar languages and {A}rabic dialect identification: A report on the third
  {DSL} shared task}.
\newblock In \emph{Proceedings of the Third Workshop on {NLP} for Similar
  Languages, Varieties and Dialects ({V}ar{D}ial3)}, pages 1--14, Osaka, Japan.
  The COLING 2016 Organizing Committee.

\bibitem[{Meftouh et~al.(2015)Meftouh, Harrat, Jamoussi, Abbas, and
  Smaili}]{Meftouh:2015:machine}
Karima Meftouh, Salima Harrat, Salma Jamoussi, Mourad Abbas, and Kamel Smaili.
  2015.
\newblock \href {https://aclanthology.org/Y15-1004} {Machine translation
  experiments on {PADIC}: A parallel {A}rabic {DI}alect corpus}.
\newblock In \emph{Proceedings of the 29th Pacific Asia Conference on Language,
  Information and Computation}, pages 26--34, Shanghai, China.

\bibitem[{Messaoudi et~al.(2022)Messaoudi, Fourati, Haddad, and
  Ben~HajHmida}]{messaoudi-2022-icompass}
Abir Messaoudi, Chayma Fourati, Hatem Haddad, and Moez Ben~HajHmida. 2022.
\newblock {iCompass Working Notes for the Nuanced Arabic Dialect Identification
  Shared task}.
\newblock In \emph{Proceedings of the Seventh Arabic Natural Language
  Processing Workshop (WANLP 2022)}. Association for Computational Linguistics.

\bibitem[{Mubarak and Darwish(2014)}]{Mubarak:2014:using}
Hamdy Mubarak and Kareem Darwish. 2014.
\newblock \href {https://doi.org/10.3115/v1/W14-3601} {Using {T}witter to
  collect a multi-dialectal corpus of {A}rabic}.
\newblock In \emph{Proceedings of the {EMNLP} 2014 Workshop on {A}rabic Natural
  Language Processing ({ANLP})}, pages 1--7, Doha, Qatar. Association for
  Computational Linguistics.

\bibitem[{Mubarak et~al.(2020)Mubarak, Darwish, Magdy, Elsayed, and
  Al-Khalifa}]{mubarak2020overview}
Hamdy Mubarak, Kareem Darwish, Walid Magdy, Tamer Elsayed, and Hend Al-Khalifa.
  2020.
\newblock \href {https://aclanthology.org/2020.osact-1.7} {Overview of {OSACT}4
  {A}rabic offensive language detection shared task}.
\newblock In \emph{Proceedings of the 4th Workshop on Open-Source Arabic
  Corpora and Processing Tools, with a Shared Task on Offensive Language
  Detection}, pages 48--52, Marseille, France. European Language Resource
  Association.

\bibitem[{Obeid et~al.(2019)Obeid, Salameh, Bouamor, and
  Habash}]{obeid-etal-2019-adida}
Ossama Obeid, Mohammad Salameh, Houda Bouamor, and Nizar Habash. 2019.
\newblock \href {https://doi.org/10.18653/v1/N19-4002} {{ADIDA}: Automatic
  dialect identification for {A}rabic}.
\newblock In \emph{Proceedings of the 2019 Conference of the North {A}merican
  Chapter of the Association for Computational Linguistics (Demonstrations)},
  pages 6--11, Minneapolis, Minnesota. Association for Computational
  Linguistics.

\bibitem[{Qaddoumi(2022)}]{qaddoumi-2022-arabic}
Abdelrahim Qaddoumi. 2022.
\newblock {Arabic Sentiment Ensemble NADI Shared Task 2}.
\newblock In \emph{Proceedings of the Seventh Arabic Natural Language
  Processing Workshop (WANLP 2022)}. Association for Computational Linguistics.

\bibitem[{Sadat et~al.(2014)Sadat, Kazemi, and Farzindar}]{sadat2014automatic}
Fatiha Sadat, Farzindar Kazemi, and Atefeh Farzindar. 2014.
\newblock \href {https://doi.org/10.3115/v1/W14-5904} {Automatic identification
  of {A}rabic language varieties and dialects in social media}.
\newblock In \emph{Proceedings of the Second Workshop on Natural Language
  Processing for Social Media ({S}ocial{NLP})}, pages 22--27, Dublin, Ireland.
  Association for Computational Linguistics and Dublin City University.

\bibitem[{Salameh et~al.(2018)Salameh, Bouamor, and
  Habash}]{Salameh:2018:fine-grained}
Mohammad Salameh, Houda Bouamor, and Nizar Habash. 2018.
\newblock \href {https://aclanthology.org/C18-1113} {Fine-grained {A}rabic
  dialect identification}.
\newblock In \emph{Proceedings of the 27th International Conference on
  Computational Linguistics}, pages 1332--1344, Santa Fe, New Mexico, USA.
  Association for Computational Linguistics.

\bibitem[{Shammary et~al.(2022)Shammary, Chen, Kardkov\'acs, Afli, and
  Alam}]{shammary-2022-tfidf}
Fouad Shammary, Yiyi Chen, Zsolt~T. Kardkov\'acs, Haithem Afli, and Mehwish
  Alam. 2022.
\newblock {TF-IDF or Transformers for Arabic Dialect Identification? ITFLOWS
  participation in the NADI 2022 Shared Task}.
\newblock In \emph{Proceedings of the Seventh Arabic Natural Language
  Processing Workshop (WANLP 2022)}. Association for Computational Linguistics.

\bibitem[{Shamsi and Abdallah(2022)}]{a2022sentiment}
Arwa A.~Al Shamsi and Sherief Abdallah. 2022.
\newblock \href {https://doi.org/10.3390/bdcc6020057} {Sentiment analysis of
  {E}mirati dialect}.
\newblock \emph{Big Data Cogn. Comput.}, 6(2):57.

\bibitem[{Sobhy et~al.(2022)Sobhy, Atta, El{-}Sawy, and
  Nayel}]{sobhy-2022-word}
Mahmoud Sobhy, Ahmed H. Abu~El Atta, Ahmed~A. El{-}Sawy, and Hamada Nayel.
  2022.
\newblock {Word Representation Models for Arabic Dialect Identification}.
\newblock In \emph{Proceedings of the Seventh Arabic Natural Language
  Processing Workshop (WANLP 2022)}. Association for Computational Linguistics.

\bibitem[{Thomas(2014)}]{thomas2014meaning}
Jenny~A Thomas. 2014.
\newblock \emph{Meaning in interaction: An introduction to pragmatics}.
\newblock Routledge.

\bibitem[{Wolf et~al.(2020)Wolf, Debut, Sanh, Chaumond, Delangue, Moi, Cistac,
  Rault, Louf, Funtowicz, Davison, Shleifer, von Platen, Ma, Jernite, Plu, Xu,
  Le~Scao, Gugger, Drame, Lhoest, and Rush}]{wolf-2020-transformers}
Thomas Wolf, Lysandre Debut, Victor Sanh, Julien Chaumond, Clement Delangue,
  Anthony Moi, Pierric Cistac, Tim Rault, Remi Louf, Morgan Funtowicz, Joe
  Davison, Sam Shleifer, Patrick von Platen, Clara Ma, Yacine Jernite, Julien
  Plu, Canwen Xu, Teven Le~Scao, Sylvain Gugger, Mariama Drame, Quentin Lhoest,
  and Alexander Rush. 2020.
\newblock \href {https://doi.org/10.18653/v1/2020.emnlp-demos.6} {Transformers:
  State-of-the-art natural language processing}.
\newblock In \emph{Proceedings of the 2020 Conference on Empirical Methods in
  Natural Language Processing: System Demonstrations}, pages 38--45, Online.
  Association for Computational Linguistics.

\bibitem[{Zaghouani and Charfi(2018)}]{Zaghouani:2018:araptweet}
Wajdi Zaghouani and Anis Charfi. 2018.
\newblock \href {https://aclanthology.org/L18-1111} {Arap-tweet: A large
  multi-dialect {T}witter corpus for gender, age and language variety
  identification}.
\newblock In \emph{Proceedings of the Eleventh International Conference on
  Language Resources and Evaluation ({LREC} 2018)}, Miyazaki, Japan. European
  Language Resources Association (ELRA).

\bibitem[{Zaidan and Callison-Burch(2011)}]{zaidan2011arabic}
Omar~F. Zaidan and Chris Callison-Burch. 2011.
\newblock \href {https://aclanthology.org/P11-2007} {The {A}rabic online
  commentary dataset: an annotated dataset of informal {A}rabic with high
  dialectal content}.
\newblock In \emph{Proceedings of the 49th Annual Meeting of the Association
  for Computational Linguistics: Human Language Technologies}, pages 37--41,
  Portland, Oregon, USA. Association for Computational Linguistics.

\bibitem[{Zampieri et~al.(2017)Zampieri, Malmasi, Ljube{\v{s}}i{\'c}, Nakov,
  Ali, Tiedemann, Scherrer, and Aepli}]{zampieri2017findings}
Marcos Zampieri, Shervin Malmasi, Nikola Ljube{\v{s}}i{\'c}, Preslav Nakov,
  Ahmed Ali, J{\"o}rg Tiedemann, Yves Scherrer, and No{\"e}mi Aepli. 2017.
\newblock \href {https://doi.org/10.18653/v1/W17-1201} {Findings of the
  {V}ar{D}ial evaluation campaign 2017}.
\newblock In \emph{Proceedings of the Fourth Workshop on {NLP} for Similar
  Languages, Varieties and Dialects ({V}ar{D}ial)}, pages 1--15, Valencia,
  Spain. Association for Computational Linguistics.

\bibitem[{Zampieri et~al.(2018)Zampieri, Malmasi, Nakov, Ali, Shon, Glass,
  Scherrer, Samard{\v{z}}i{\'c}, Ljube{\v{s}}i{\'c}, Tiedemann, van~der Lee,
  Grondelaers, Oostdijk, Speelman, van~den Bosch, Kumar, Lahiri, and
  Jain}]{zampieri2018language}
Marcos Zampieri, Shervin Malmasi, Preslav Nakov, Ahmed Ali, Suwon Shon, James
  Glass, Yves Scherrer, Tanja Samard{\v{z}}i{\'c}, Nikola Ljube{\v{s}}i{\'c},
  J{\"o}rg Tiedemann, Chris van~der Lee, Stefan Grondelaers, Nelleke Oostdijk,
  Dirk Speelman, Antal van~den Bosch, Ritesh Kumar, Bornini Lahiri, and Mayank
  Jain. 2018.
\newblock \href {https://aclanthology.org/W18-3901} {Language identification
  and morphosyntactic tagging: The second {V}ar{D}ial evaluation campaign}.
\newblock In \emph{Proceedings of the Fifth Workshop on {NLP} for Similar
  Languages, Varieties and Dialects ({V}ar{D}ial 2018)}, pages 1--17, Santa Fe,
  New Mexico, USA. Association for Computational Linguistics.

\bibitem[{Zhang et~al.(2022)Zhang, Abdul-Mageed, and Nagoudi}]{zhang2022decay}
Chiyu Zhang, Muhammad Abdul-Mageed, and El~Moatez~Billah Nagoudi. 2022.
\newblock \href {https://doi.org/10.36190/2022.92} {Decay no more: A persistent
  twitter dataset for learning social meaning}.
\newblock \emph{Workshop Proceedings of the 16th International AAAI Conference
  on Web and Social Media}.

\bibitem[{Zitouni et~al.(2020)Zitouni, Abdul-Mageed, Bouamor, Bougares, El-Haj,
  Tomeh, and Zaghouani}]{wanlp-2020-arabic}
Imed Zitouni, Muhammad Abdul-Mageed, Houda Bouamor, Fethi Bougares, Mahmoud
  El-Haj, Nadi Tomeh, and Wajdi Zaghouani, editors. 2020.
\newblock \href {https://aclanthology.org/2020.wanlp-1.0} {\emph{Proceedings of
  the Fifth Arabic Natural Language Processing Workshop}}. Association for
  Computational Linguistics, Barcelona, Spain (Online).

\end{thebibliography}
